\documentclass{article}

\usepackage{microtype}
\usepackage{graphicx}
\usepackage{subfigure}
\usepackage{wrapfig}
\usepackage{booktabs} 

\usepackage{hyperref}

\usepackage[accepted]{icml2021}

\usepackage{amsmath}
\usepackage{amssymb}
\usepackage{mathrsfs}
\usepackage{amsthm}
\usepackage{stmaryrd}
\usepackage{mathtools}

\usepackage{bm}
\usepackage{dblfloatfix}

\usepackage{multirow}
\usepackage{xcolor}

\newcommand{\cA}{\mathcal{A}}

\newcommand{\cP}{\mathcal{P}}
\newcommand{\cR}{\mathcal{R}}
\newcommand{\cS}{\mathcal{S}}

\newcommand{\rE}{\mathbb{E}}

\newcommand{\rR}{\mathbb{R}}

\icmltitlerunning{Revisiting Rainbow: Promoting more Insightful and Inclusive Deep Reinforcement Learning Research}

\begin{document}

\twocolumn[
\icmltitle{Revisiting Rainbow: Promoting more Insightful and Inclusive Deep Reinforcement Learning Research}



\icmlsetsymbol{equal}{*}

\begin{icmlauthorlist}
\icmlauthor{Johan S. Obando-Ceron}{}
\icmlauthor{Pablo Samuel Castro}{goo}
\end{icmlauthorlist}

\icmlaffiliation{goo}{Google Research, Brain Team}

\icmlcorrespondingauthor{Johan S. Obando-Ceron}{jobando0730@gmail.com}
\icmlcorrespondingauthor{Pablo Samuel Castro}{psc@google.com}

\icmlkeywords{Reinforcement Learning, MDPs}

\vskip 0.3in
]

\printAffiliationsAndNotice{}

\begin{abstract}
  Since the introduction of DQN, a vast majority of reinforcement learning research has focused on reinforcement learning with deep neural networks as function approximators. New methods are typically evaluated on a set of environments that have now become standard, such as Atari 2600 games. While these benchmarks help standardize evaluation, their computational cost has the unfortunate side effect of widening the gap between those with ample access to computational resources, and those without. In this work we argue that, despite the community's emphasis on large-scale environments, the traditional small-scale environments can still yield valuable scientific insights and can help reduce the barriers to entry for underprivileged communities. To substantiate our claims, we empirically revisit the paper which introduced the Rainbow algorithm \citep{hessel18rainbow} and present some new insights into the algorithms used by Rainbow.
\end{abstract}

\section{Introduction}
Since the introduction of DQN \citep{mnih2015humanlevel} reinforcement learning has witnessed a dramatic increase in research papers \citep{henderson2017reinforcement}. A large portion of these papers propose new methods that build on the original DQN algorithm and network architecture, often adapting methods introduced before DQN to work well with deep networks (e.g., \citep{hasselt2015doubledqn,bacon17option,castro20scalable}). New methods are typically evaluated on a set of environments that have now become standard, such as the Atari 2600 games made available in the Arcade Learning Environment (ALE) \citep{bellemare2012ale} and the control tasks available in MuJoCo and DM control suites \citep{todorov12mujoco,tassa2020dmcontrol}.

While these benchmarks have helped to evaluate new methods in a standardized manner, they have also implicitly established a minimum amount of computing power in order to be recognized as valid scientific contributions. Although classic reinforcement learning tasks such as MountainCar, CartPole, Acrobot, and grid worlds have not gone away, they are now used mostly for evaluating theoretical contributions (e.g., \citep{nachum2019dualdice,lyle19comparative}); indeed, in our experience it is quite difficult to have a paper proposing a new reinforcement learning method accepted at one of the major machine learning conferences unless it includes experiments with one of the benchmarks just mentioned. This is unfortunate, as the low computational cost and speed at which one can train on small-scale environments enables broad hyper-parameter sweeps and more thorough investigations into the nuances of the methods being considered, as well as the reevaluation of certain empirical choices that have become ``folk wisdom'' in these types of experiments.

Furthermore, at a time when efforts such as Black in AI and LatinX in AI are helping bring people from underrepresented (and typically underprivileged) segments of society into the research community, these newcomers are faced with enormous computational hurdles to overcome if they wish to be an integral part of said community.

It thus behooves the reinforcement learning research community to incorporate a certain degree of flexibility and creativity when proposing and evaluating new research; of course, this should not be at the expense of scientific rigour. This paper is partly a position paper, partly an empirical evaluation. We argue for a need to change the status-quo in evaluating and proposing new research to avoid exacerbating the barriers to entry for newcomers from underprivileged communities. In \autoref{sec:revisitingRainbow} we complement this argument by revisiting the Rainbow algorithm \citep{hessel18rainbow}, which proposed a new state of the art algorithm by combining a number of recent advances, on a set of small- and medium-sized tasks. This allows us to conduct a ``counterfactual'' analysis: would \citet{hessel18rainbow} have reached the same conclusions if they had run on the smaller-scale experiments we investigate here? In \autoref{sec:beyondTheRainbow} we extend this analysis by investigating the interaction between the different algorithms considered and the network architecture used, varying the distribution parameterization and bootstrapping methodology, and the base loss used by DQN. Finally, in \autoref{sec:putting} we compare the Rainbow variants we considered, as well as provide insights into the properties of the different environments used in our study.

\section{Preliminaries}
Reinforcement learning methods are used for learning how to act (near) optimally in sequential decision making problems in uncertain environments. In the most common scenario, an {\em agent} transitions between {\em states} in an environment by making {\em action} choices at discrete time steps; upon performing an action, the environment produces a numerical {\em reward} and transitions the agent to a new state. This is formalized as a Markov decision process \citep{puterman94mdp} $\langle \cS,\cA,\cR,\cP,\gamma\rangle$, where $\cS$ is a finite set of states, $\cA$ is a finite set of actions, $\cR:\cS\times\cA\rightarrow [R_{min}, R_{max}]$ is the reward function, $\cP:\cS\times\cA\rightarrow\Delta(\cS)$ is the transition function, where $\Delta(X)$ denotes the set of probability distributions over a set $X$, and $\gamma\in[0, 1)$ is a horizon discount factor.

An agent's behaviour is formalized as a policy $\pi:\cS\rightarrow\Delta(\cA)$ which induces a {\em value function} $V^{\pi}:\cS\rightarrow\rR$:
\begin{align*}
  V^{\pi}(s) & := \rE \left[ \sum_{t=0}^{\infty} \gamma^t \cR(s_t, a_t) | s_0=s, s_t, a_t\right] \\
  & = \rE_{a\sim\pi(s)} \left[ \cR(s, a) + \gamma \rE_{s'\sim\cP(s, a)} V^{\pi}(s') \right],
\end{align*}

where $s_t\sim \cP(s_{t-1}, a_{t-1})$ and $a_t\sim\pi(s_t)$. The second line is the well-known Bellman recurrence. It is also convenient to consider the value of actions that differ from those encoded in $\pi$ via the function $Q^{\pi}:\cS\times\cA\rightarrow\rR$: $Q^{\pi}(s, a) := \cR(s, a) + \gamma \rE_{s'\sim\cP(s, a)} V^{\pi}(s')$.

It is well known that there exist policies $\pi^*$ that are optimal in the sense that $V^*(s) := V^{\pi^*}(s) \geq V^{\pi}(s)$ for all policies $\pi$ and states $s\in\cS$. In reinforcement learning we are typically interested in having agents find these policies by interacting with the environment. One of the most popular ways to do so is via $Q$-learning, where an agent maintains a function $Q_\theta$, parameterized by $\theta$ (e.g. the weights in a neural network), and updates it after observing the transition $s\overset{a,r}{\rightarrow}s'$ using the method of {\em temporal differences} \citep{sutton88learning}:
\begin{align}
  Q_{\theta}(s, a) \leftarrow Q_{\theta}(s, a) + \alpha \left( Q^*(s', a') - Q_{\theta}(s, a)\right)
  \label{eqn:qlearning}
\end{align}
where $\alpha$ is a learning rate and the optimal target values $ Q^*(s', a') = r + \gamma\max_{a'\in\cA}Q_{\theta}(s', a')$

\subsection{DQN}
\citet{mnih2015humanlevel} introduced DQN, which combined Q-learning with deep networks. Some of the most salient design choices are:
\begin{itemize}
  \item The $Q$ function is represented using a feed forward neural network consisting of three convolutional layers followed by two fully connected layers. Two copies of the $Q$-network are maintained: an {\em online} network (parameterized by $\theta$) and a {\em target} network (parameterized by $\bar{\theta}$). The online network is updated  via the learning process described below, while the target network remains fixed and is synced with the online weights at less frequent (but regular) intervals.
  \item A large {\em replay buffer} $D$ is maintained to store experienced transitions $(s, a, r, s')$ \citep{lin92self}.
  \item The update in \autoref{eqn:qlearning} is implemented using the following {\em loss function} to update the {\em online} network:
    \begin{align}
    L(\theta) = \rE_{(s, a, r, s')\sim U(D)}[\left( Y^{DQN} - Q_{\theta}(s, a)\right)^2]
    \label{eqn:dqnloss}
    \end{align}
    using $Y^{DQN} = \left(r + \gamma\max_{a'\in\cA}Q_{\bar{\theta}}(s', a')\right)$ and {\em mini-batches} of size 32.
\end{itemize}

\subsection{Rainbow}
In this section we briefly present the enhancements to DQN that were combined by \citet{hessel18rainbow} for the Rainbow agent.

{\bf Double Q-learning: } \citet{hasselt2015doubledqn} added double Q-learning \citep{hasselt10double} to mitigate overestimation bias in the $Q$-estimates by decoupling the maximization of the action from its selection in the target bootstrap.
%

{\bf Prioritized experience replay: } Instead of sampling uniformly from the replay buffer ($U(D)$), prioritized experience replay \citep{schaul2015prioritized} proposed to sample a trajectory $t=(s, a, r, s')$ with probability $p_t$ proportional to the temporal difference error.

{\bf Dueling networks: } \citet{wang16dueling} introduced dueling networks by modifying the DQN network architecture. Specifically, two streams share the initial convolutional layers, separately estimating $V^*(s)$, and the advantages for each action: $A(s, a) := Q^*(s, a) - V^*(s)$. The output of the network is a combination of these two streams.
%
%
%
{\bf Multi-step learning: } Instead of computing the temporal difference error using a single-step transition, one can use multi-step targets instead \cite{sutton88learning}, where for a trajectory $(s_0, a_0, r_0, s_1, a_1, \cdots)$ and update horizon $n$: $R_t^{(n)} := \sum_{k=0}^{n-1}\gamma^k r_{t+k+1}$, yielding the multi-step temporal difference: $R_t^{(n)} + \gamma^n\max_{a'\in\cA}Q_{\bar{\theta}}(s_{t+n}, a') - Q_{\theta}(s_t, a_t)$.

{\bf Distributional RL: } \citet{bellemare17distributional} demonstrated that the Bellman recurrence also holds for {\em value distributions}: $Z(x, a) \overset{D}{=} R(s, a) + \gamma Z(X', A')$, where $Z$, $R$, and $(X', A')$ are random variables representing the return, immediate reward, and next state-action, respectively. The authors present an algorithm (C51) to maintain an estimate $Z_{\theta}$ of the return distribution $Z$ by use of a parameterized categorical distribution with 51 atoms.

{\bf Noisy nets: } \citet{fortunato18noisy} propose replacing the simple $\epsilon$-greedy exploration strategy used by DQN with noisy linear layers that include a noisy stream.

\begin{figure*}[!t]
  \centering
   \includegraphics[width=\textwidth]{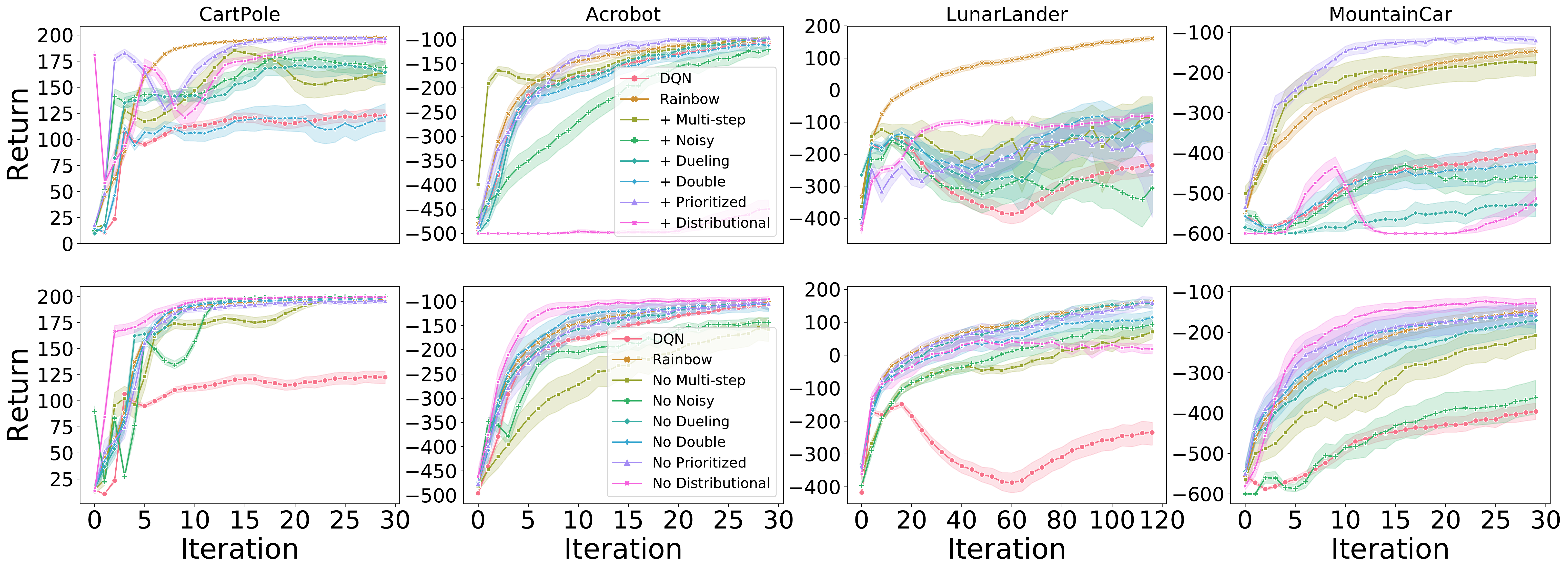}
   \vspace{-0.5cm}
  \caption{Comparison of the different algorithmic components on the four small environments using the optimal hyper-parameters from \autoref{tbl:classicHyperParams} for each, averaged over 100 independent runs (shaded areas show 95\% confidence intervals). Top row explores adding on top of DQN, bottom row explores removing from Rainbow.}
  \label{fig:revisitingClassic}
\end{figure*}

\section{The cost of Rainbow}
\label{sec:costOfRainbow}
Although the {\em value} of the hybrid agent uncovered by \citet{hessel18rainbow} is undeniable, this result could have only come from a large research laboratory with ample access to compute.
Indeed, each Atari 2600 game from the ALE (there are 57 in total) takes roughly 5 days to fully train using specialized hardware (for example, an NVIDIA Tesla P100 GPU).\footnote{The computational expense is not limited to the ALE: MuJoCo tasks from the DM control suite take about 2 days with the same hardware.} Additionally, in order to be able to report performance with confidence bounds it is common to use at least five independent runs \citep{henderson2017reinforcement,machado18revisiting}.

Thus, to provide the convincing empirical evidence for Rainbow, \citet{hessel18rainbow} required approximately 34,200 GPU hours (or 1425 days). Note that this cost does not include the hyper-parameter tuning that was necessary to optimize the various components. Considering that the cost of a Tesla P100 GPU is around US\$6,000, providing this evidence will take an unreasonably long time as it is prohibitively expensive to have multiple GPUs in a typical academic lab so they can be used in parallel. As a reference point, the average minimum monthly wage in South America (excluding Venezuela) is approximately US\$313\footnote{Taken from https://www.statista.com/statistics/953880/latin-america-minimum-monthly-wages/}; in other words, one GPU is the equivalent of approximately 20 minimum wages. Needless to say, this expectation is far from inclusive.

In light of this, we wish to investigate three questions: 
\begin{enumerate}
  \item Would \citet{hessel18rainbow} have arrived at the same qualitative conclusions, had they run their experiments on a set of smaller-scale experiments?
  \item Do the results of \citet{hessel18rainbow} generalize well to non-ALE environments, or are their results overly-specific to the chosen benchmark?
  \item Is there scientific value in conducting empirical research in reinforcement learning when restricting oneself to small- to mid-scale environments?
\end{enumerate}
We investigate the first two in Section \ref{sec:revisitingRainbow}, and the last in Sections \ref{sec:beyondTheRainbow} and \ref{sec:putting}.

\section{Revisiting Rainbow}
\label{sec:revisitingRainbow}
\subsection{Methodology}
We follow a similar process as \citet{hessel18rainbow} in evaluating the various algorithmic variants mentioned above: we investigate the effect of adding each on top of the original DQN agent as well as the effect of dropping each from the final Rainbow agent, sweeping over learning rates for each. Our implementation is based on the Dopamine framework \citep{castro18dopamine}. Note that Dopamine includes a ``lite'' version of Rainbow, which does not include noisy networks, double DQN, nor dueling networks, but we have added all these components in our implementation\footnote{Source code available at\\https://github.com/JohanSamir/revisiting\_rainbow}.

\begin{figure*}[!t]
  \centering
  \includegraphics[width=\textwidth]{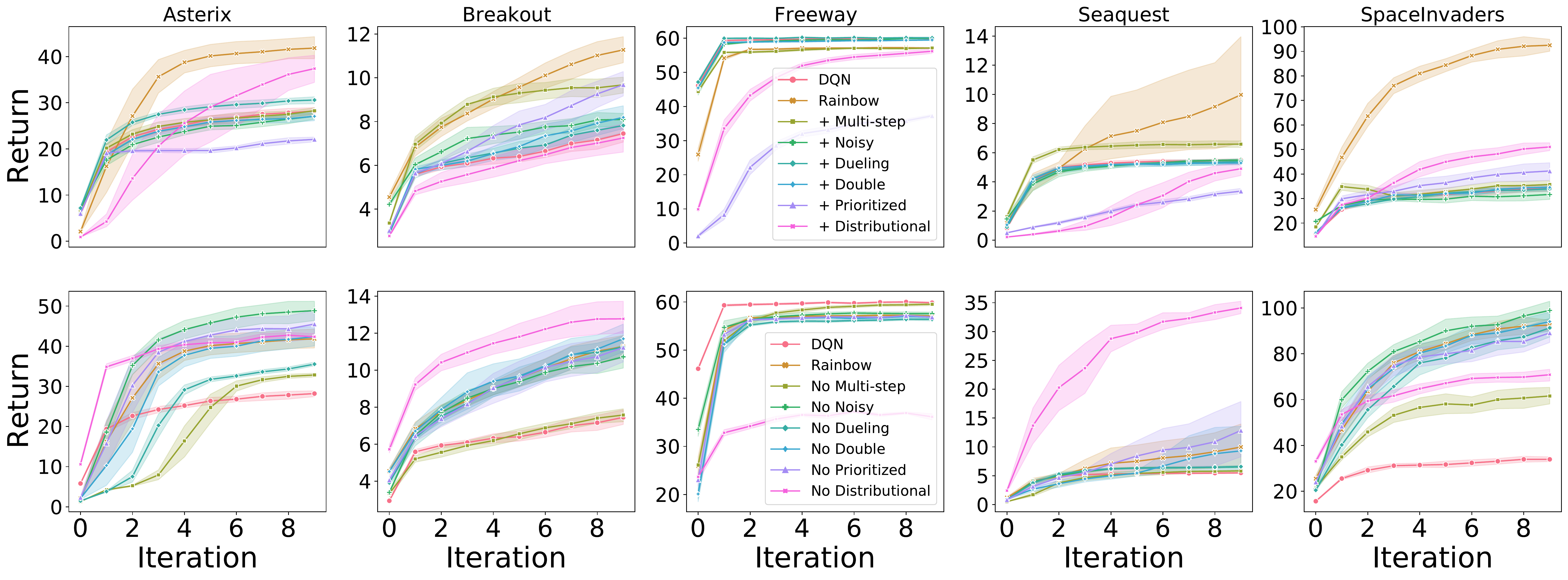}
   \vspace{-0.5cm}
  \caption{Comparison of the different algorithmic components on the five MinAtar games, averaged over 10 independent runs (shaded areas show 90\% confidence intervals). Top row explores adding on top of DQN, bottom row explores removing from Rainbow.}
  \label{fig:revisitingMinatar}
\end{figure*}

We perform our empirical evaluation on small-scale environments (CartPole, Acrobot, LunarLander, and MountainCar) which are all available as part of the OpenAI Gym library \citep{brockman16gym} (see \autoref{sec:environments} for a detailed description of each environment). We used multilayer perceptrons (MLPs) with 2 layers of 512 units each for these experiments. The agents were all trained on a CPU; it is worth noting that of these environments the one that took longest to train (LunarLander) is still able to finish in under two hours.

In order to strengthen the Rainbow Connection, we also ran a set of experiments on the MinAtar environment \citep{young19minatar}, which is a set of miniaturized versions of five ALE games (Asterix, Breakout, Freeway, Seaquest, and SpaceInvaders). These environments are considerably larger than the four classic control environments previously explored, but they are significantly faster to train than regular ALE environments. Specifically, training one of these agents takes approximately 12-14 hours on a P100 GPU.  For these experiments, we followed the network architecture used by \citet{young19minatar} consisting of a single convolutional layer followed by a linear layer.

\subsection{Empirical evaluation}

Under constant hyper-parameter settings (see appendix for details), we evaluate both the {\em addition} of each algorithmic component to DQN, as well as their {\em removal} from the full Rainbow agent on the classic control environments (\autoref{fig:revisitingClassic}) and on Minatar (\autoref{fig:revisitingMinatar}).

We analyze our results in the context of the first two questions posed in Section~\ref{sec:costOfRainbow}: {\em Would \citet{hessel18rainbow} have arrived at the same qualitative conclusions, had they run their experiments on a set of smaller-scale experiments? Do the results of \citet{hessel18rainbow} generalize well to non-ALE environments, or are their results overly-specific to the chosen benchmark?}

What we find is that the performance of the different components is not uniform throughout all environments; a finding which is consistent with the results observed by \citet{hessel18rainbow}. However, if we were to suggest a single agent that balances the tradeoffs of the different algorithmic components, our analysis would be consistent with \citet{hessel18rainbow}: combining all components produces a better overall agent.

Nevertheless, there are important details in the variations of the different algorithmic components that merit a more thorough investigation. An important finding of our work is that distributional RL, when added on its own to DQN, may actually hurt performance (e.g. Acrobot and Freeway); similarly, performance can sometimes increase when distributional RL is removed from Rainbow (e.g. MountainCar and Seaquest); this is in contrast to what was found by \citet{hessel18rainbow} on the ALE experiments and warrants further investigation. As \citet{lyle19comparative} noted, under non-linear function approximators (as we are using in these experiments), using distributional RL generally produces {\em different} outcomes than the non-distributional variant, but these differences are not always positive.

\begin{figure*}[!h]
  \centering
  \includegraphics[width=\textwidth]{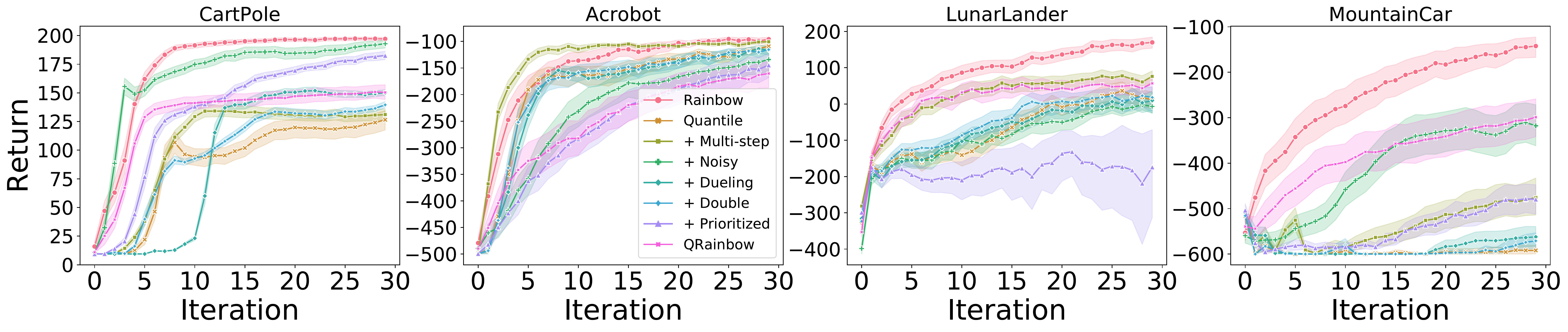}
  \includegraphics[width=\textwidth]{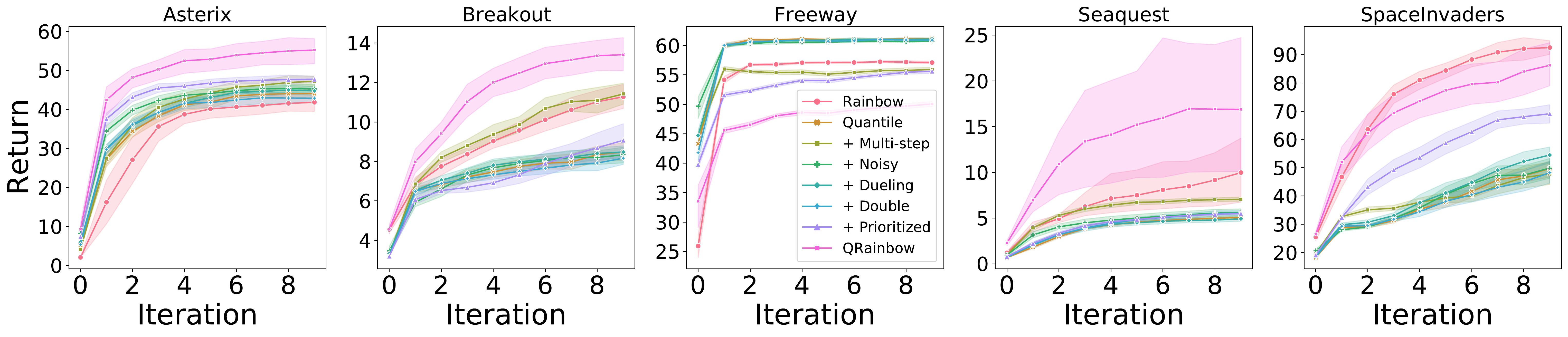}
   \vspace{-0.5cm}
  \caption{Investigating the combination of the different algorithmic components with QR-DQN (shaded areas show 95\% (top) and 90\% (bottom) confidence intervals).}
  \label{fig:comparisonQuantile}
\end{figure*}

\section{Beyond the Rainbow}
\label{sec:beyondTheRainbow}

In this section we seek to answer the third question posed in \autoref{sec:costOfRainbow}: {\em Is there scientific value in conducting empirical research in small- to mid-scale environments?} We leverage the low cost of the small-scale environments to conduct a thorough investigation into some of the algorithmic components studied. Unless otherwise specified, the classic control and MinAtar environments are averaged over 100 and 10 independent runs, respectively; in both cases shaded areas report 95\% confidence intervals.

\subsection{Examining network architectures and batch sizes}
We investigated the interaction of the best per-game hyper-parameters with the number of layers and units per layer. Due to space constraints we omit the figures from the main paper, but include them in the appendix. We found that in general using 2-3 layers with at least 256 units each yielded the best performance. Further, aside from Cartpole, the algorithms were generally robust to varying network architecture dimensions.

Another often overlooked hyper-parameter in training RL agents is the batch size. We investigated the sensitivity of DQN and Rainbow to varying batch sizes and found that while for DQN it is sub-optimal to use a batch size below 64, Rainbow seems fairly robust to both small and large batch sizes.

\subsection{Examining distribution parameterizations}
Although distributional RL is an important component of the Rainbow agent, at the time of its development Rainbow was only evaluated with the C51 parameterization of the distribution, as originally proposed by \citet{bellemare17distributional}. 
Since then there have been a few new proposals for parameterizing the return distribution, notably quantile regression \citep{dabney2017distributional,dabney18distributional} and implicit quantile networks \citep{dabney18iqn}. In this section we investigate the interaction of these parameterizations with the other Rainbow components.

{\bf Quantile Regression for Distributional RL}

In contrast to C51, QR-DQN \cite{dabney2017distributional,dabney18distributional} computes the return quantile values for $N$ fixed, uniform probabilities. Compared to C51, QR-DQN has no restrictions or bound for value, as the distribution of the random return is approximated by a uniform mixture of $N$ Diracs: $Z_{\theta}(x, a):=\frac{1}{N} \sum_{i=1}^{N} \delta_{\theta_{i}(x, a)}$,
with each $\theta_{i}$ assigned a quantile value trained with quantile regression. In \autoref{fig:comparisonQuantile} we evaluate the interaction of the different Rainbow components with Quantile and find that, in general, QR-DQN responds favourably when augmented with each of the components. We also evaluate a new agent, {\em QRainbow}, which is the same as Rainbow but with the QR-DQN parameterization. It is interesting to observe that in the classic control environments Rainbow outperforms QRainbow, but QRainbow tends to perform better than Rainbow on Minatar (with the notable exception of Freeway), suggesting that perhaps the quantile parameterization of the return distribution has greater benefits when used with networks that include convolutional layers.

\begin{figure*}[!h]
  \centering
  \includegraphics[width=\textwidth]{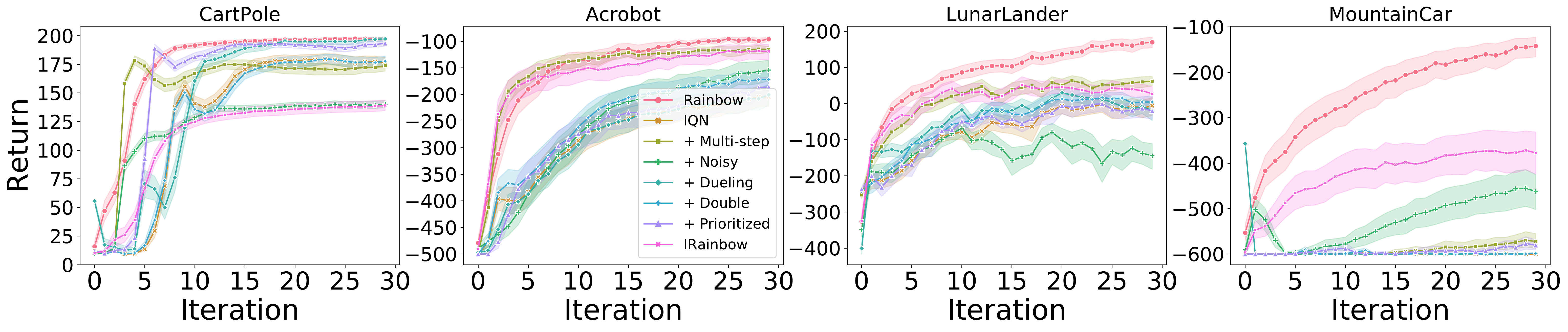}
  \includegraphics[width=\textwidth]{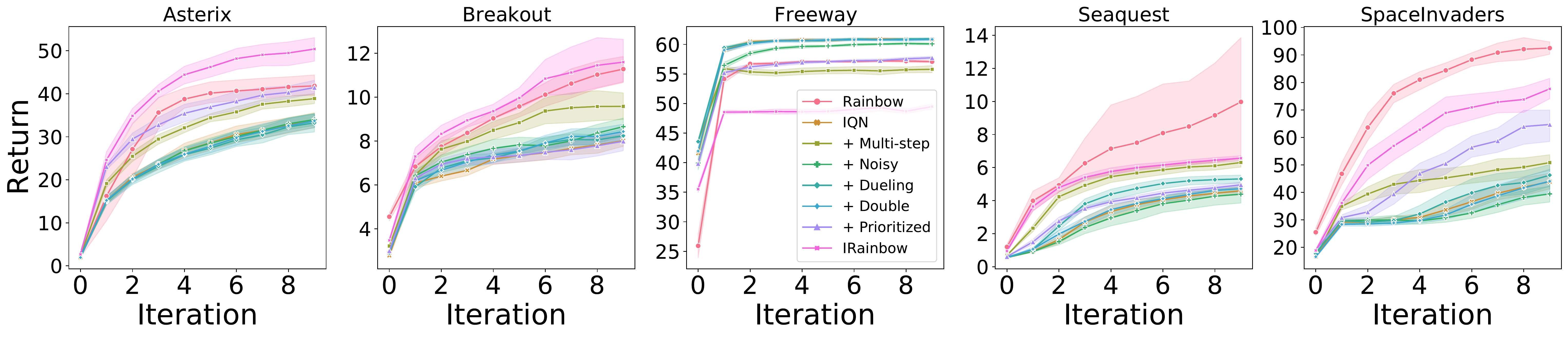}
   \vspace{-0.5cm}
  \caption{Investigating the combination of the different algorithmic components with IQN (shaded areas show 95\% (top) and 90\% (bottom) confidence intervals).}
  \label{fig:comparisonIQN}
\end{figure*}

{\bf Implicit quantile networks}

We continue by investigating using implicit quantile networks (IQN) as the parameterization of the return distribution \citet{dabney18iqn}. IQN learns to transform a base distribution (typically a uniform distribution in $[0, 1]$) to the quantile values of the return distribution. This can result in greater representation power in comparison to QR-DQN, as well as the ability to incorporate {\em distortion risk measures}. We repeat the ``Rainbow experiment'' with IQN and report the results in \autoref{fig:comparisonIQN}. In contrast to QR-DQN, in the classic control environments the effect on performance of various Rainbow components is rather mixed and, as with QR-DQN {\em IRainbow} underperforms Rainbow. In Minatar we observe a similar trend as with QR-DQN: IRainbow outperforms Rainbow on all the games except Freeway.

\subsection{Munchausen Reinforcement Learning}

\citet{vieillard2020munchausen} introduced {\em Munchausen RL} as a simple variant to any temporal difference learning agent consisting of two main components: the use of stochastic policies and augmenting the reward with the scaled log-policy. Integrating their proposal to DQN yields M-DQN with performance superior to that of C51; the integration of Munchausen-RL to IQN produced M-IQN, a new state-of-the art agent on the ALE.

\begin{figure*}[!h]
  \centering
  \includegraphics[width=\textwidth]{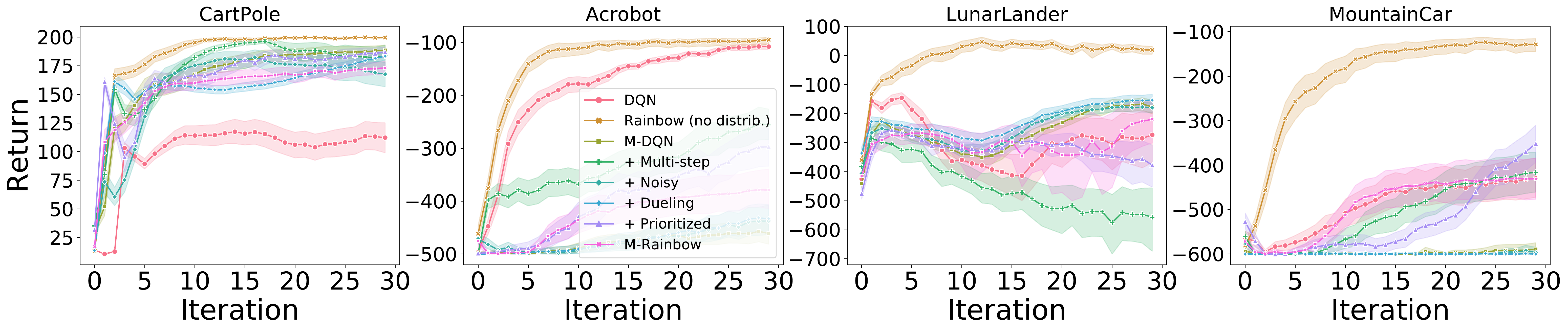}
  \includegraphics[width=\textwidth]{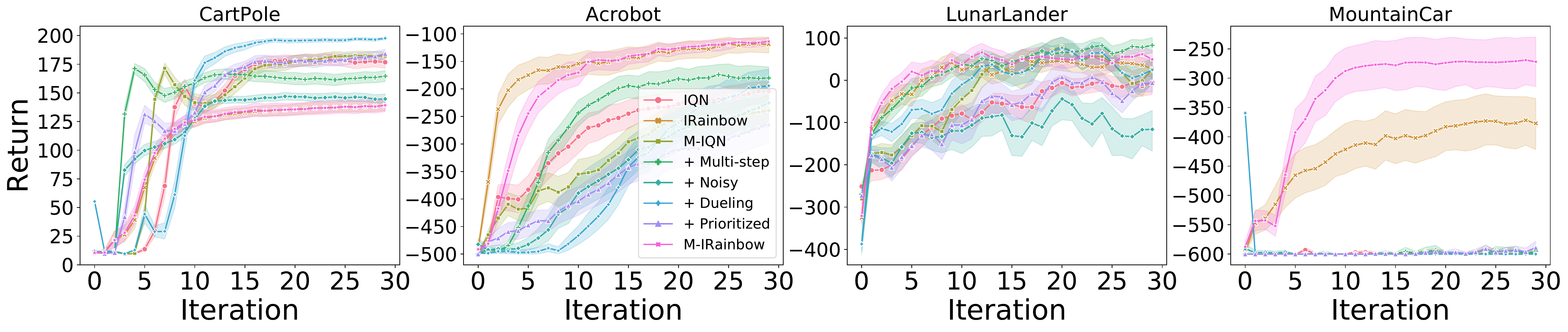}
  \includegraphics[width=\textwidth]{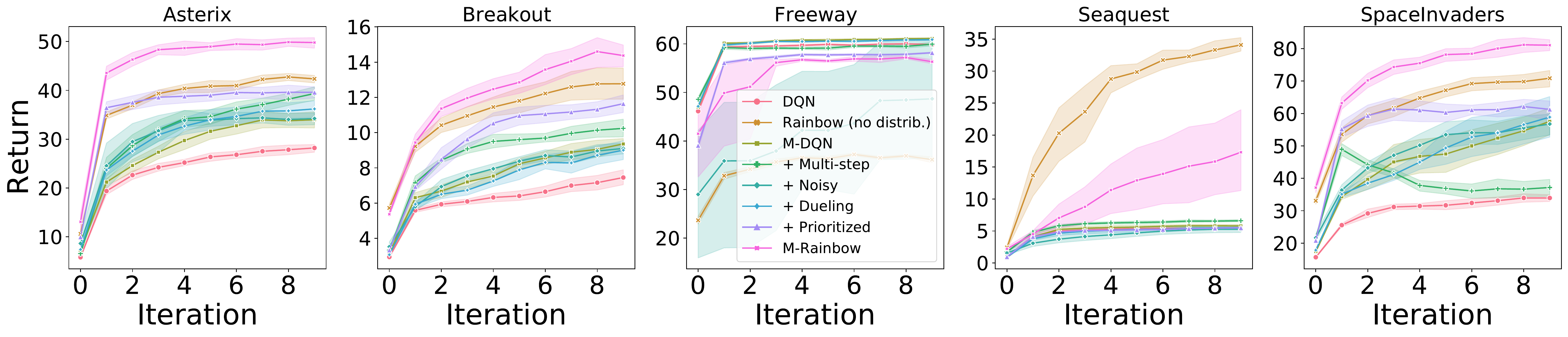}
  \includegraphics[width=\textwidth]{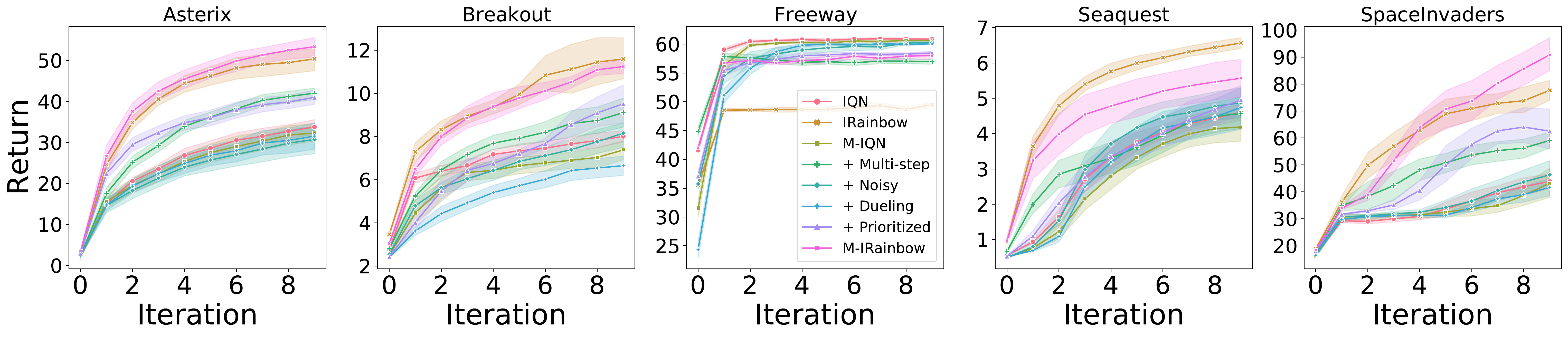}
   \vspace{-0.5cm}
  \caption{Investigating the combination of the different algorithmic components with M-DQN and M-IQN (shaded areas show 95\% (top 2) and 90\% (bottom 2) confidence intervals).}.
  \label{fig:comparisonMunchausen}
\end{figure*}

In \autoref{fig:comparisonMunchausen} we report the results when repeating the Rainbow experiment on M-DQN and M-IQN.
In the classic control environments neither of the Munchausen variants seem to yield much of an improvement over their base agents. In Minatar, while M-DQN does seem to improve over DQN, the same cannot be said of M-IQN. We explored combining all the Rainbow components\footnote{We were unable to successfully integrate M-DQN with C51 nor double-DQN, so our M-Rainbow agent is compared against Rainbow without distributional RL and without double-DQN.} with the Munchausen agents and found that, in the classic control environments, while M-Rainbow underperforms relative to its non-Munchausen counterpart, M-IRainbow can provide gains. In Minatar, the results vary from game to game, but it appears that the Munchausen agents yield an advantage on the same games (Asterix, Breakout, and SpaceInvaders).

\subsection{Reevaluating the Huber loss}
The Huber loss is what is usually used to train DQN agents as it is meant to be less sensitive to outliers. Based on recent anecdotal evidence, we decided to evaluate training DQN using the mean-squared error (MSE) loss and found the surprising result that on {\em all} environments considered using the MSE loss yielded much better results than using the Huber loss, sometimes even surpassing the performance of the full Rainbow agent (full classic control and Minatar results are provided in the appendix). This begs the question as to whether the Huber loss is truly the best loss to use for DQN, especially considering that reward clipping is typically used for most ALE experiments, mitigating the occurence of outlier observations. Given that we used Adam \citep{kingma15adam} for all our experiments while the original DQN algorithm used RMSProp, it is important to consider the choice of optimizer in answering this question.

\begin{figure*}[!h]
  \centering
  \includegraphics[width=\textwidth]{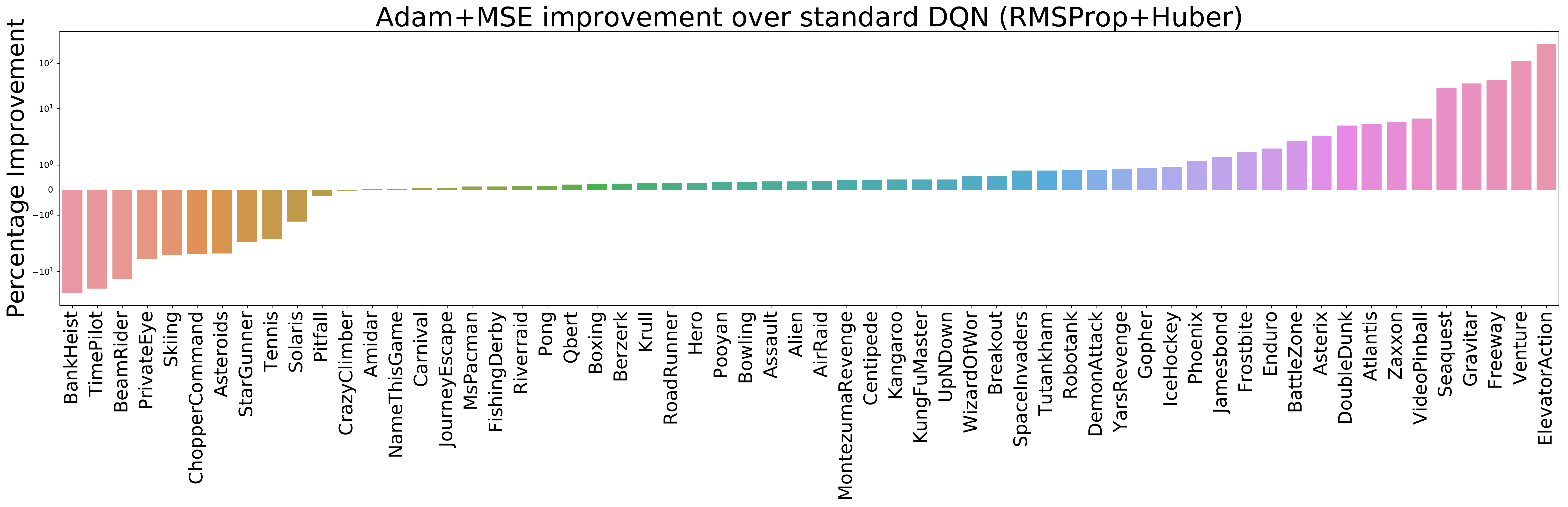}
   \vspace{-0.5cm}
  \caption{Comparison of training DQN using the Adam optimizer with the MSE loss, versus the Huber loss with the RMSProp. All results report the average of 5 independent runs.}
  \label{fig:adamMSEVsHuber}
\end{figure*}

To obtain an answer, we compared the performance of the Huber versus the MSE loss when used with both the Adam and RMSProp optimizers on all 60 Atari 2600 games. In \autoref{fig:adamMSEVsHuber} we present the improvement obtained when using the Adam optimizer with the MSE loss over using the RMSProp optimizer with the Huber loss and find that, overwhelmingly, Adam+MSE is a superior combination than RMSProp+Huber. In the appendix we provide complete comparisons of the various optimizer-loss combinations that confirm our finding. Our analyses also show that, when using RMSProp, the Huber loss tends to perform better than MSE, which in retrospect explains why \cite{mnih2015humanlevel} chose the Huber over the simpler MSE loss when introducing DQN.

Our findings highlight the importance in properly evaluating the interaction of the various components used when training RL agents, as was also argued by \citet{fujimoto20equivalence} with regards to loss functions and non-uniform sampling from the replay buffer; as well as by \citet{hessel19inductive} with regards to inductive biases used in training RL agents.

\section{Putting it all together}
\label{sec:putting}

\subsection{Rainbow flavours}
We compare the performance of DQN against all of the Rainbow variants and show the results of two environments in \autoref{fig:aggregateComparisons} (full comparisons in the appendix). These two environments highlight the fact that, although Rainbow does outperform DQN, there are important differences amongst the various flavours that invite further investigation.

\begin{figure}[!h]
  \centering
  \includegraphics[width=0.5\textwidth]{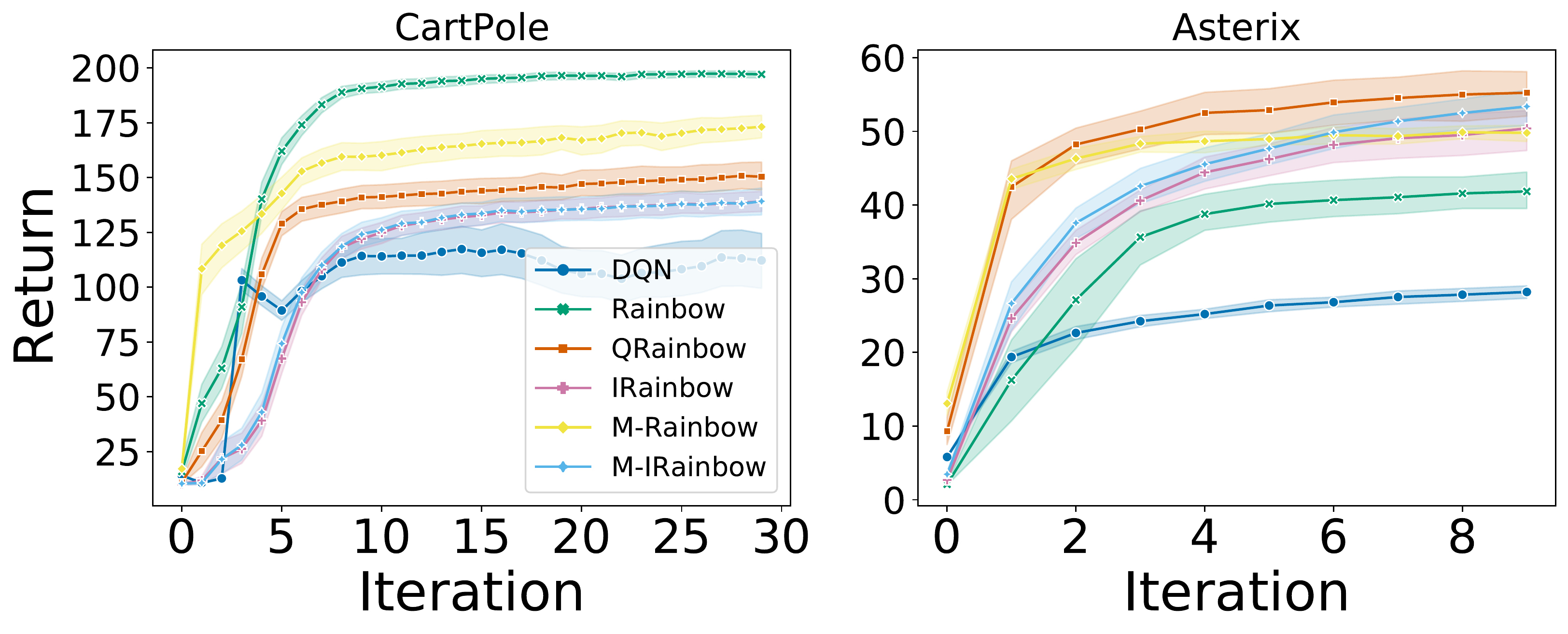}
   \vspace{-0.5cm}
  \caption{Comparing DQN against the various Rainbow flavours.}
  \label{fig:aggregateComparisons}
\end{figure}

\subsection{Environment properties}

Our exhaustive experimentation on the four classic control and five MinAtar environments grant us some insight into their differing properties. We believe these environments pose a variety of interesting challenges for RL research and present a summary of our insights here, with a more thorough analysis in the appendix, in the hope that it may stimulate future investigations.

{\bf Classic control}

Although {\bf CartPole} is a relatively simple task, DQN can be quite sensitive to learning rates and as such, can prove an interesting testbed for optimizer stability. We found the use of Noisy networks and MSE loss to dramatically help with this sensitivity. It seems that distributional RL is required for obtaining good performance on {\bf LunarLander} (see the poor results with DQN, M-DQN, and M-Rainbow in comparison with the others), suggesting this would be a good environment to investigate the differences in expectational and distributional RL as proposed by \citet{lyle19comparative}. Both {\bf Acrobot} and {\bf MountainCar} are sparse reward environments, which are typically good environments for investigating exploration strategies; indeed, we observe that Noisy networks tend to give a performance gain in both these environments. MountainCar appears to be the more difficult of the two, a fact also observed by \citet{tang17exploration}, \citet{colas18gep}, and \citet{declan2020analyzing}.

{\bf MinAtar}

The value of MinAtar environments has already been argued by \citet{young19minatar} and recently exemplified by \citet{ghiassian20gradient} and \citet{buckman2021importance}. It is worth highlighting that both {\bf Seaquest} and {\bf Freeway} appear to lend themselves well for research on exploration methods, due to their partial observability and reward sparsity. We would like to stress that these environments enable researchers to investigate the importance and effect of using convolutional networks in RL without the prohibitive expense of the ALE benchmark.

\section{Conclusion}
On a limited computational budget we were able to reproduce, at a high-level, the findings of \citet{hessel18rainbow} and uncover new and interesting phenomena. Evidently it is much easier to revisit something than to discover it in the first place; however, our intent with this work was to argue for the relevance and significance of empirical research on small- and medium-scale environments. We believe that these less computationally intensive environments lend themselves well to a more critical and thorough analysis of the performance, behaviours, and intricacies of new algorithms (a point also argued by \citet{osband2020bsuite}). It is worth remarking that when we initially ran 10 independent trials for the classic control environments, the confidence intervals were very wide and inconclusive; boosting the independent trials to 100 gave us tighter confidence intervals with small amounts of extra compute. This would be impractical for most large-scale environments. Ensuring statistical significance when comparing algorithms, a point made by \citet{colas18how} and \citet{jordan20evaluating}, is facilitated with the ability to run a large number of independent trials.

We are by no means calling for less emphasis to be placed on large-scale benchmarks. We are simply urging researchers to consider smaller-scale environments as a valuable tool in their investigations, and reviewers to avoid dismissing empirical work that focuses on these smaller problems. By doing so, we believe, we will get both a clearer picture of the research landscape and will reduce the barriers for newcomers from diverse, and often underprivileged, communities. These two points can only help make our community and our scientific advances stronger.

\section{Acknowledgements}
The authors would like to thank Marlos C. Machado, Sara Hooker, Matthieu Geist, Nino Vieillard, Hado van Hasselt, Eleni Triantafillou, and Brian Tanner for their insightful comments on our work.

\bibliographystyle{plainnat}
\bibliography{revisiting_rainbow}

\clearpage
\onecolumn

\appendix

\section*{\centering APPENDICES: Revisiting Rainbow}

\section{Environments}
\label{sec:environments}
For OpenAI environments, data is summarized from \href{https://github.com/openai/gym}{https://github.com/openai/gym} and information provided on the wiki \href{https://github.com/openai/gym/wiki}{https://github.com/openai/gym/wiki}.

\begin{figure}[!h]
  \centering
  \includegraphics[width=0.23\textwidth]{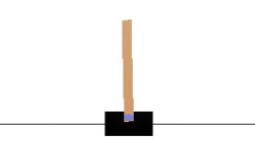}
  \includegraphics[width=0.23\textwidth]{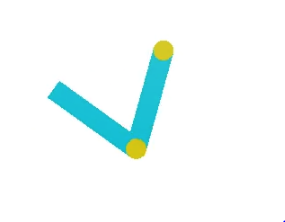}
  \includegraphics[width=0.28\textwidth]{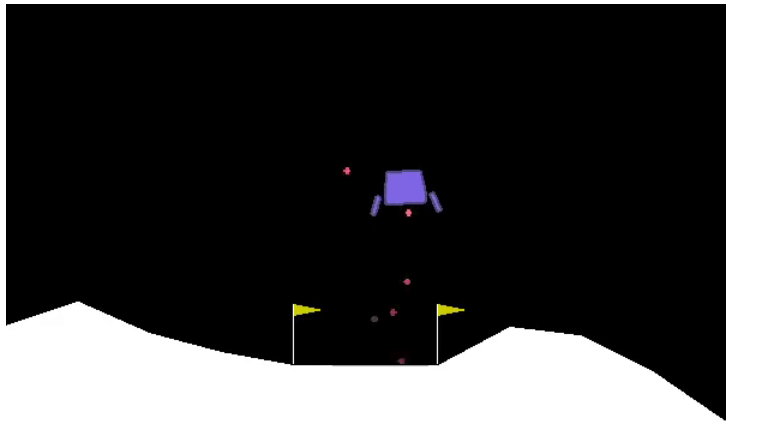}
  \includegraphics[width=0.22\textwidth]{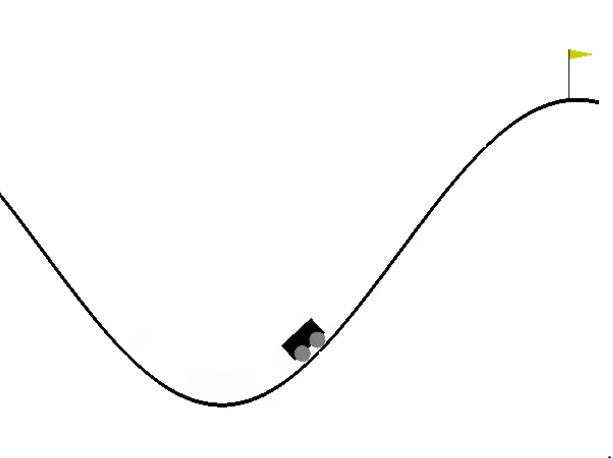}
  \caption{The classic control environments. From left to right: CartPole, Acrobot, LunarLander, and MountainCar.}
\end{figure}

\subsection{CartPole-v0}
CartPole is a task of balancing a pole on top of the cart. The cart has access to its position and velocity as state, and can only go left or right for each action. The task is over when the pole falls over (less than $\pm 12$ deg), the cart goes out of the boundaries ($\pm 2.4$ units off the center), or 200 time steps are reached, with each step returning 1 reward. The agent is given a continuous 4-dimensional space describing the environment, and can respond by returning one of two values, pushing the cart either right or left.

\subsection{Acrobot-v1}
In the Acrobot environment, the agent is given rewards for swinging a double-jointed pendulum up from a stationary position. The agent can actuate the second joint by returning one of three actions, corresponding to left, right, or no torque. The agent is given a six dimensional vector describing the environment's angles and velocities. The episode ends when the end of the second pole is more than the length of a pole above the base. For each timestep that the agent does not reach this state, it is given a -1 reward. The episode length is 500 timesteps.

\subsection{LunarLander-v2}
In the LunarLander environment, the agent attempts to land a lander on a particular location on a simulated 2D world. If the lander hits the ground going too fast, the lander will explode, or if the lander runs out of fuel, the lander will plummet toward the surface. The agent is given a continuous vector describing the state, and can turn its engine on or off. The landing pad is placed in the center of the screen, and if the lander lands on the pad, it is given a reward (100-140 points). The agent also receives a variable amount of reward when coming to rest, or contacting the ground with a leg (10 points). The agent loses a small amount of reward by firing the engine (-0.3 points), and loses a large amount of reward if it crashes (-100 points).  The observation consists of the x and y coordinates, the x and y velocities, angle, angular velocity, and ground contact information of the lander (left and right leg) and the action consists of do nothing, fire left orientation engine, fire down engine and fire right orientation engine.

\subsection{MountainCar-v0}
MountainCar is a one dimensional track between two mountains. The goal is to drive up the mountain to the right. The agent receives a -1 reward for every time step it does  not reach the top. The episode terminates when it reaches 0.5 position, or if 200 iterations are reached. The objective is for the agent to learn to drive back and forth to build momentum that will be enough to push the car up the hill. The observation consists of the car’s position and velocity and the action consists of pushing left, pushing right and no push.

\subsection{MinAtar Environments}
Some of the best reinforcement learning algorithms require tens or hundreds of millions of timesteps to  learn to play Atari games, the equivalent of several weeks of training in real time. MinAtar reduces the complexity of the representation learning problem for 5 Atari games, while maintaining the mechanics of the original games as much as possible. We use these MinAtar games to measure the impact of using some extensions to the DQN algorithm. 
MinAtar provides analogues to five Atari games which play out on a 10x10 grid. The environments provide a 10x10xn state representation, where each of the n channels correspond to a game-specific object, such as ball, paddle and brick in the game Breakout. Detailed descriptions of each game are available in \href{https://github.com/kenjyoung/MinAtar}{https://github.com/kenjyoung/MinAtar}.

\begin{figure}[h]
    \centering
    \includegraphics[width=0.8\textwidth]{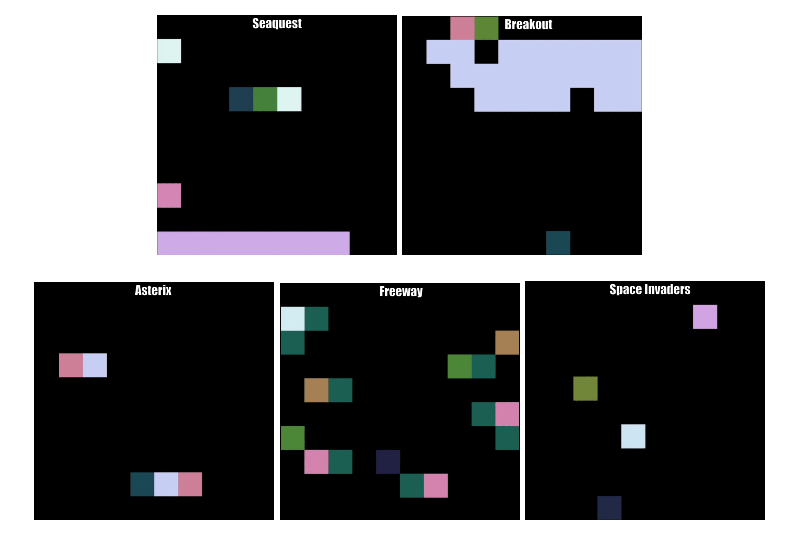}
    \caption{Visualization of each MinAtar environment. }
    \label{fig:mesh1}
\end{figure}
\clearpage

\section{Detailed description of the revisting rainbow components}
\label{sec:description_components}

In this section we present the enhancements to DQN that were combined  for the revisting rainbow agent.

\subsection{Double Q-learning}
 \citet{hasselt2015doubledqn} added double Q-learning \citep{hasselt10double} to mitigate overestimation bias in the $Q$-estimates by decoupling the maximization of the action from its selection in the target bootstrap. The loss from \autoref{eqn:dqnloss} is replaced with
\begin{align*}
  L(\theta) = \rE_{(s, a, r, s')\sim U(D)}\left[\left( r + \gamma Q_{\bar{\theta}}(s', \arg\max_{a'\in\cA}Q_{\theta}(s', a')) - Q_{\theta}(s, a)\right)^2\right]
\end{align*}

\subsection{Prioritized experience replay}
Instead of sampling uniformly from the replay buffer ($U(D)$), prioritized experience replay \citep{schaul2015prioritized} proposed to sample a trajectory $t=(s, a, r, s')$ with probability $p_t$ proportional to the temporal difference error:
\[ p_t \propto \left| r + \gamma \max_{a'\in\cA}Q_{\bar{\theta}}(s', a') - Q_{\theta}(s, a)\right|^{\omega} \]
where $\omega$ is a hyper-parameter for the sampling distribution.

\subsection{Dueling networks}
\citet{wang16dueling} introduced dueling networks by modifying the DQN network architecture. Specifically, two streams share the initial convolutional layers, separately estimating $V^*(s)$, and the advantages for each action: $A(s, a) := Q^*(s, a) - V^*(s)$.
The output of the full network is defined by:
\[ Q_{\theta}(s, a) = V_{\eta}(f_{\eta}(s)) + A_{\psi}(f_{\eta}(s), a) - \frac{\sum_{a'\in\cA} A_{\psi}(f_{\xi}(s), a')}{|\cA|} \]
Where $\xi$ denotes the parameters for the shared convolutional layers, $\eta$ denotes the parameters for the value estimator stream, $\psi$ denots the parameters for the advantage estimator stream, and $\theta := \xi\cup\eta\cup\psi$.

\subsection{Multi-step learning}
 Instead of computing the temporal difference error using a single-step transition, one can use multi-step targets instead \cite{sutton88learning}, where for a trajectory $(s_0, a_0, r_0, s_1, a_1, \cdots)$ and update horizon $n$:
\[ R_t^{(n)} := \sum_{k=0}^{n-1}\gamma^k r_{t+k+1} \]
yielding the multi-step temporal difference:
\[ R_t^{(n)} + \gamma^n\max_{a'\in\cA}Q_{\bar{\theta}}(s_{t+n}, a') - Q_{\theta}(s_t, a_t) \]

\subsection{Distributional RL}
\citet{bellemare17distributional} demonstrated that the Bellman recurrence also holds for {\em value distributions}:
\[ Z(x, a) \overset{D}{=} R(s, a) + \gamma Z(X', A') \]
where $Z$, $R$, and $(X', A')$ are random variables representing the return, immediate reward, and next state-action, respectively. The authors present an algorithm (C51) to maintain an estimate $Z_{\theta}$ of the return distribution $Z$ by use of a parameterized categorical distribution with 51 atoms.

\subsection{Quantile Regression for Distributional RL}

\citet{dabney2017distributional} computes the return quantile values for $N$ fixed, uniform probabilities.  The distribution of the random return is approximated by a uniform mixture of $N$ Diracs
with each $\theta_{i}$ assigned a quantile value trained with quantile regression. Formally, a quantile distribution $Z_{\theta} \in \mathcal{Z}_{Q}$ maps each state-action pair $(x, a)$ to a uniform probability distribution is,
\[Z_{\theta}(x, a):=\frac{1}{N} \sum_{i=1}^{N} \delta_{\theta_{i}(x, a)} \]
These quantile estimates are trained using the Huber \citep{robusthuber} quantile regression loss.

\subsection{Implicit quantile networks}

\citet{dabney18iqn} extend  the  approach  of  Dabney  et  al.(2018), from learning a discrete set of quantiles to learningthe full quantile function, a continuous map from probabilities to returns. Moreover, extended the fixed quantile fractions to uniform samples.  The approximation of the quantile function $Z$ is,

\[Z_{\tau}(x, a) \approx f(\psi(x) \odot \phi(\tau))_{a}\]

where $\odot$ denotes the element-wise (Hadamard) product.

\subsection{Noisy nets}
 \citet{fortunato18noisy} propose replacing the simple $\epsilon$-greedy exploration strategy used by DQN with noisy linear layers that include a noisy stream. Specifically, the standard linear layers defined by $\mathbf{y} = \mathbf{b} + \mathbf{Wx}$ are replaced by:
\[ \mathbf{y} = (\mathbf{b} + \mathbf{Wx}) + (\mathbf{b_{noisy}}\odot\epsilon^b + (\mathbf{W_{noisy}}\odot\epsilon^w)\mathbf{x}) \]
where $\epsilon^b$ and $\epsilon^w$ are noise variables (\citet{hessel18rainbow} use factorised Gaussian noise)  and $\odot$ is the Hadamard product.

\subsection{Munchausen Reinforcement Learning}
\citet{vieillard2020munchausen} modify the regression target adding the scaled log-policy to the immediate reward: 
\[Y^{M-DQN}=r_{t}+\alpha \tau \ln \pi_{\bar{\theta}}\left(a_{t} \mid s_{t}\right)+\gamma \sum_{a^{\prime} \in \mathcal{A}} \pi_{\bar{\theta}}\left(a^{\prime} \mid s_{t+1}\right) \left(Q_{\bar{\theta}}\left(s_{t+1}, a^{\prime}\right)-\tau \ln \pi_{\bar{\theta}}\left(a^{\prime} \mid s_{t+1}\right)\right) \]

with $\pi_{\theta}=\operatorname{sm}\left(\frac{Q_{\bar{\theta}}}{\tau}\right)$ and a scaling factor $\alpha \in[0,1]$.

\subsection{Loss functions}

We ran  experiments using Mean Square Error and Huber loss \cite{robusthuber}, therefore we consider it important to introduce these concepts in this paper. MSE is the sum of squared distances between the target variable and predicted values. MSE gives relatively higher weight (penalty) to large errors/outliers, while smoothening the gradient for smaller errors. The MSE is formally defined by the following equation,
\[\mathrm{MSE}(y_i, \hat{y}_i)= \frac{1}{2} \left(y_{i}-\hat{y}_{i}\right)^{2} \]

On the other hand,  Huber loss  is less sensitive to outliers in data than the squared error loss and it is defined by the following equation,

\[ \mathrm{Huber}\left(y_{i}, \hat{y}_{i}\right)=\left\{\begin{array}{lr}\frac{1}{2}\left(y_{i}-\hat{y}_{i}\right)^{2} & \text { for }\left|y_{i}-\hat{y}_{i}\right| \leq \delta, \\  \delta\left|y_{i}-\hat{y}_{i}\right|-\frac{1}{2}\delta^{2} & \text { otherwise }\end{array}\right.\]

\clearpage

\section{Hyperparameters settings for classic control environments}
\label{sec:parameters}

\begin{table}[!h]
 \centering
  \caption{Hyperparameters settings for classic control environments}
  \label{tbl:classicHyperParams}
 \begin{tabular}{@{} ccccccc @{}}
    \toprule
   Hyperparameter & DQN &  Rainbow & QR-DQN & IQN & M-DQN & M-IQN \\
    \midrule
   gamma & 0.99 & 0.99 &  0.99 &  0.99 &  0.99 & 0.99 \\
   update horizon  & 1 & 3 & 1 & 1 & 1 &1 \\
   min replay history  & 500 & 500 & 500 & 500 & 500 &500 \\
   update period & 4 & 2 & 2 & 2 & 4 & 2 \\
   target update period  & 100 & 100 & 100 & 100 & 100 &100 \\
   normalize obs  & True & True & True & True & True & True \\
   hidden layer & 2 & 2 & 2 & 2 & 2 & 2\\
   neurons & 512 & 512 & 512 & 512 & 512 & 512 \\
   \midrule
   num atoms  & - & 51 & 51 & - & - &- \\
   vmax  & -  & 200  &-  &-  &-  &-  \\
   kappa & -  & -  &1  &1  &-  &1 \\
   \midrule
   tau & - & - & - & - & 100 & 0.03 \\
   alpha   & -  & - & - & - & 1 & 1\\
   clip value min & - & - & - & - &  -1e3 &-1 \\
   \midrule
   num tau samples & - & - & - & 32 & - & 32 \\
   num tau prime samples  & - & - & - & 32 & - & 32 \\
   num quantile samples & - & - & - & 32 & - & 32 \\
   quantile embedding dim & - & - & - & 64 & - & 64 \\
   \midrule
   learning rate  & 0.001 & 0.001 & 0.001 & 0.001 & 0.001 & 0.001 \\
   eps   & 3.125e-4 & 3.125e-4 & 3.125e-4 & 3.125e-4 & 3.125e-4 & 3.125e-4 \\
   num iterations & 30 & 30 & 30 & 30 & 30 & 30 \\
   replay capacity  &  50000 &  50000 &  50000 &  50000 &  50000 & 50000\\
   batch size & 128 & 128 & 128 & 128 & 128 & 128 \\
     \bottomrule
  \end{tabular}
\end{table}

\clearpage

\section{Hyperparameters settings for MinAtar environments}
\label{sec:parametersII}

\begin{table}[!h]
 \centering
  \caption{Hyperparameters settings for MinAtar environments}
  \label{tbl:minatarHyperParams}
 \begin{tabular}{@{} ccccccc @{}}
    \toprule
   Hyperparameter & DQN &  Rainbow & QR-DQN & IQN & M-DQN & M-IQN \\
    \midrule
   gamma & 0.99 & 0.99 &  0.99 &  0.99 &  0.99 & 0.99 \\
   update horizon  & 1 & 3 & 1 & 1 & 1 &1 \\
   min replay history  & 1000 & 1000 & 1000 & 1000 & 1000 &1000 \\
   update period & 4 & 4 & 4 & 4 & 4 & 4 \\
   target update period  & 1000 & 1000 & 1000 & 1000 & 1000 &1000 \\
   normalize obs  & True & True & True & True & True & True \\
   \midrule
   num atoms  & - & 51 & 51 & - & - &- \\
   vmax  & -  & 100  &-  &-  &-  &-  \\
   kappa & -  & -  &1  &1  &-  &1 \\
   \midrule
   tau & - & - & - & - & 0.03 & 0.03 \\
   alpha   & -  & - & - & - & 0.9 & 0.9\\
   clip value min & - & - & - & - &  - 1&-1 \\
   \midrule
   num tau samples & - & - & - & 32 & - & 32 \\
   num tau prime samples  & - & - & - & 32 & - & 32 \\
   num quantile samples & - & - & - & 32 & - & 32 \\
   quantile embedding dim & - & - & - & 64 & - & 64 \\
   \midrule
   learning rate  & 0.00025 & 0.00025 & 0.00025 & 0.00025 & 0.00025 & 0.00025 \\
   eps   & 3.125e-4 & 3.125e-4 & 3.125e-4 & 3.125e-4 & 3.125e-4 & 3.125e-4 \\
   num iterations & 10 & 10 & 10 & 10 & 10 & 10 \\
   replay capacity  &  100000 &  100000 &  100000 &  100000 &  100000 & 100000\\
   batch size & 32 & 32 & 32 & 32 & 32 & 32 \\
     \bottomrule
  \end{tabular}
\end{table}

\section{Average runtime for environments used in the experiments}
\label{sec:parametersII}

\begin{table}[!h]
 \centering
  \caption{Average runtime.}
  \label{tbl:averageTime}
 \begin{tabular}{@{} cc @{}}
    \toprule
   Figures & Average time (minutes)  \\
    \midrule     
Cartpole & 10  \\
Acrobot & 10   \\
LunarLander & 30   \\
MountainCar & 10  \\
MinAtar games & 720  \\
Atari games & 7200  \\
     \bottomrule
  \end{tabular}
\end{table}

%
%
%

\clearpage
\section{Impact of algorithmic components}
In the following tables we list the algorithmic components that had the most positive effect on the various agents and environments. The following abbreviations are used in this section: Dist: Distributional, Noi: Noisy, Prio: Prioritized,
Mult: Multi-step, Dou: Double , Due: Dueling.

\begin{table}[!h]
  \centering
  \caption{Impact of algorithmic components for each agent on classic control environments.}
  \label{tbl:classicControlComp}
  \begin{tabular}{@{} cccccc @{}}
    \toprule
     Agent & Operation & Cartpole & Acrobot & LunarLander & MountainCar \\

     \midrule
	     DQN & + &  Dist/Noi  & Prio/ Mult & Dist/ Mult &  Mult/Dou  \\	       
	\midrule
	    Rainbow & - &  Mult/Noi & Mult/Noi & Dist/Mult & Mult/Noi \\
	\midrule
	     QR-DQN & + & Noi/Prio & Mult&Mult &Noi/Mult \\	
	\midrule
	    IQN & + & Prio /Due &Mult/Noi   & Mult/Dou &Noi \\                       
	\midrule
	     M-DQN & + &Mult &Mult /Prio   & Due/Noi & Mult  \\	                          
	\midrule
	    M-IQN & + &  Prio& Mult &   Mult /Due & - \\

     \bottomrule
  \end{tabular}
\end{table}

\begin{table}[!h]
  \centering
  \caption{Impact of algorithmic components for each agent on MinAtar games.}
  \label{tbl:minatarComponents}
  \begin{tabular}{@{} ccccccc @{}}
    \toprule
     Agent & Operation & Asterix& Breakout & Freeway & Seaquest & SpaceInvaders \\
     \midrule
    DQN & + &  Dist/Due & Mult/Dou &Due/Noi &Mult/Noi & Dist/ Mult \\

\midrule
      Rainbow & - & Dist/Mult & Mult/ Dist & Dist/Due& Mult/Due & Dist/Mult\\

\midrule
	QR-DQN & + & Prio/ Multi-step &Mult/ Prio& Noi  &Mult/Noi &Prio/Due \\

\midrule
     IQN & + & Prio/Noi &Mult/Noi & Due & Mult/Due & Prio/Mult \\
\midrule
     M-DQN& + &  Prio/Mult &  Prio/Mult &- & Mult & Prio/Due \\
\midrule
    M-IQN & + & Mult/Prio &  Prio/ Mult & - & Prio/Noi & Prio/ Mult \\ 
                         
     \bottomrule
  \end{tabular}
\end{table}

\begin{table}[!h]
  \centering
  \caption{ The most important component for classic control environments and MinAtar games.}
  \label{tbl:classicControlComp}
  \begin{tabular}{@{} cccccc @{}}
    \toprule
     Agent & Operation & Classic control & MinAtar \\

     \midrule
	     DQN & + &  Mult & Mult  \\	       
	\midrule
	    Rainbow & - &  Mult &  Dist \\
	\midrule
	     QR-DQN & + &Mult & Prio \\	
	\midrule
	    IQN & + & Mult & Mult \\                       
	\midrule
	     M-DQN & + & Mult &Prio \\	                          
	\midrule
	    M-IQN & + &  Mult & Prio  \\

     \bottomrule
  \end{tabular}
\end{table}

\begin{table}[!h]
  \centering
  \caption{The most important component for each environment.}
  \label{tbl:classicControlComp}
  \begin{tabular}{@{} cccccc @{}}
    \toprule
     Environment & Component \\

     \midrule
	     Cartpole &  Prio  \\	       
	\midrule
	    Acrobot & Mult  \\
	\midrule
	    LunarLander & Mult \\	
	\midrule
	    MountainCar & Noi  \\                       
	\midrule
	    Asterix & Prio    \\	                          
	\midrule
	   Breakout & Mult  \\
	\midrule
	    Freeway & Due  \\
	\midrule
	    Seaquest & Mult \\	
	\midrule
	   SpaceInvaders &  Mult \\                       
	            
     \bottomrule
  \end{tabular}
\end{table}

\clearpage

\section{Examining network architectures and batch sizes}
\begin{figure*}[!h]
  \centering
  \includegraphics[width=\textwidth]{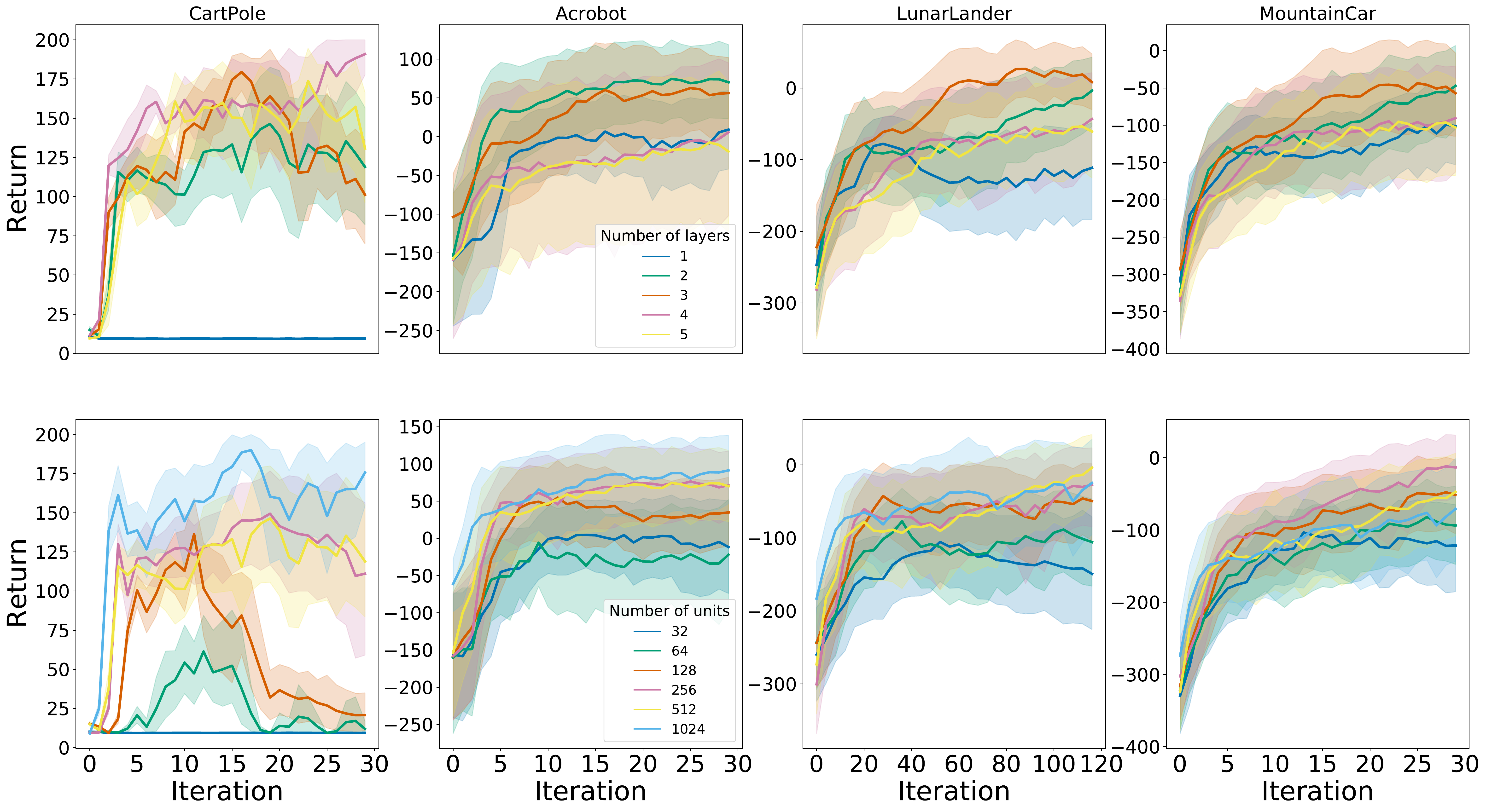}
  \caption{Evaluating DQN sensitivity to varying number of layers (top) and units per layer (bottom).}
  \label{fig:comparisonArchSweepDQN}
\end{figure*}

\begin{figure*}[!h]
  \centering
  \includegraphics[width=\textwidth]{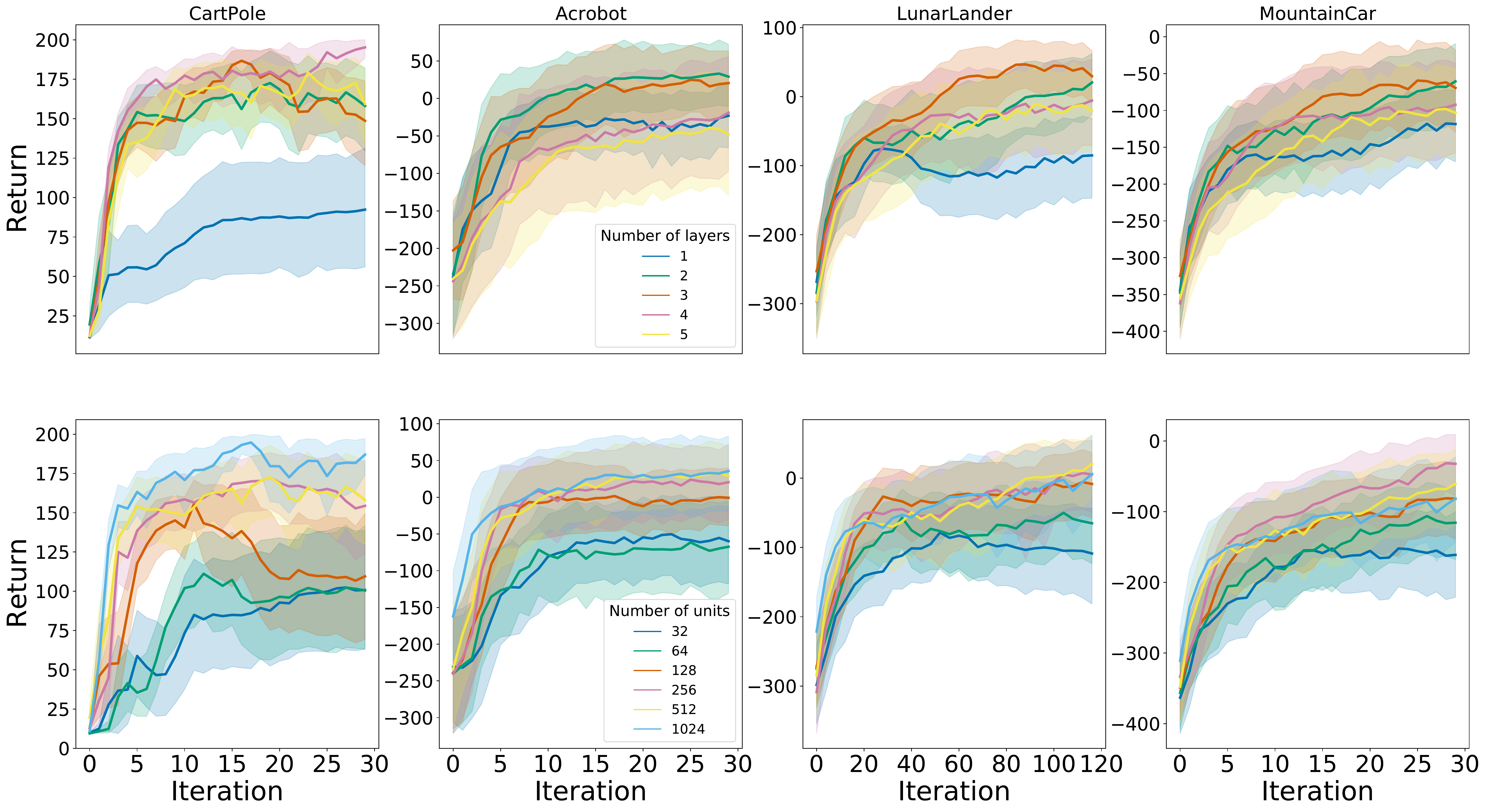}
  \caption{Evaluating Rainbow sensitivity to varying number of layers (top) and units per layer (bottom).}
  \label{fig:comparisonArchSweepRainbow}
\end{figure*}

\begin{figure*}[!h]
  \centering
  \includegraphics[width=\textwidth]{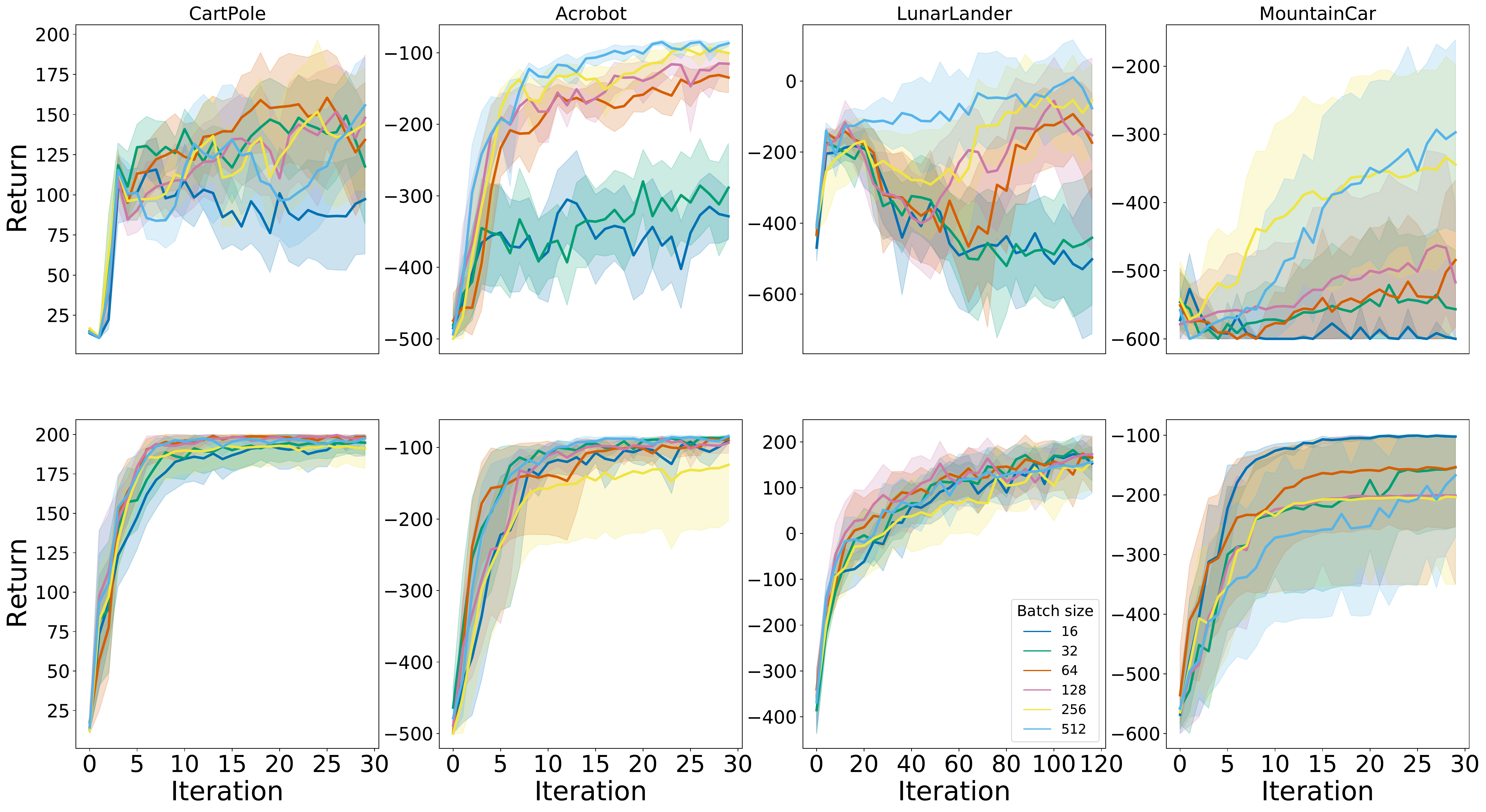}
  \caption{Evaluating DQN (top) and Rainbow (bottom) sensitivity to varying batch sizes. The default value used in the rest of the experiments is 128.}
  \label{fig:comparisonBatchSizeSweep}
\end{figure*}

In this section we evaluate the effect of varying the number of layers and number of hidden units with DQN (\autoref{fig:comparisonArchSweepDQN}) and Rainbow (\autoref{fig:comparisonArchSweepRainbow}), as well as the effect of batch sizes on both (\autoref{fig:comparisonBatchSizeSweep}). For all the results in this section we ran 10 independent runs for each, and the shaded areas represent 95\% confidence intervals.

\clearpage

\section{Reevaluating the Huber loss, complete results}

In this section we present complete results comparing the various combinations possible when using either Adam or RMSProp optimizers, and Huber or MSE losses.

\subsection{Classic control and MinAtar results}
\begin{figure*}[!h]
  \centering
  \includegraphics[width=0.24\textwidth]{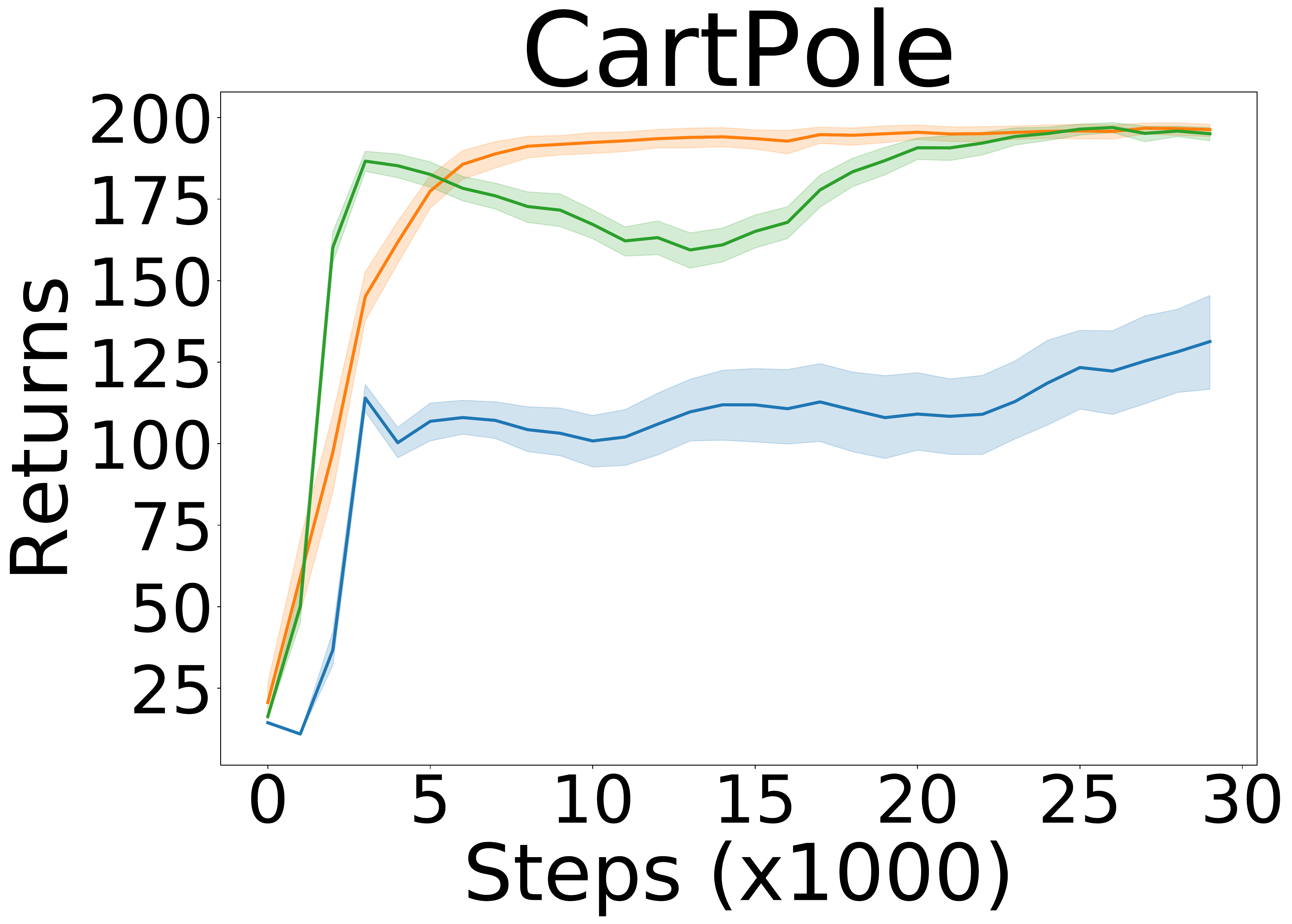}
  \includegraphics[width=0.24\textwidth]{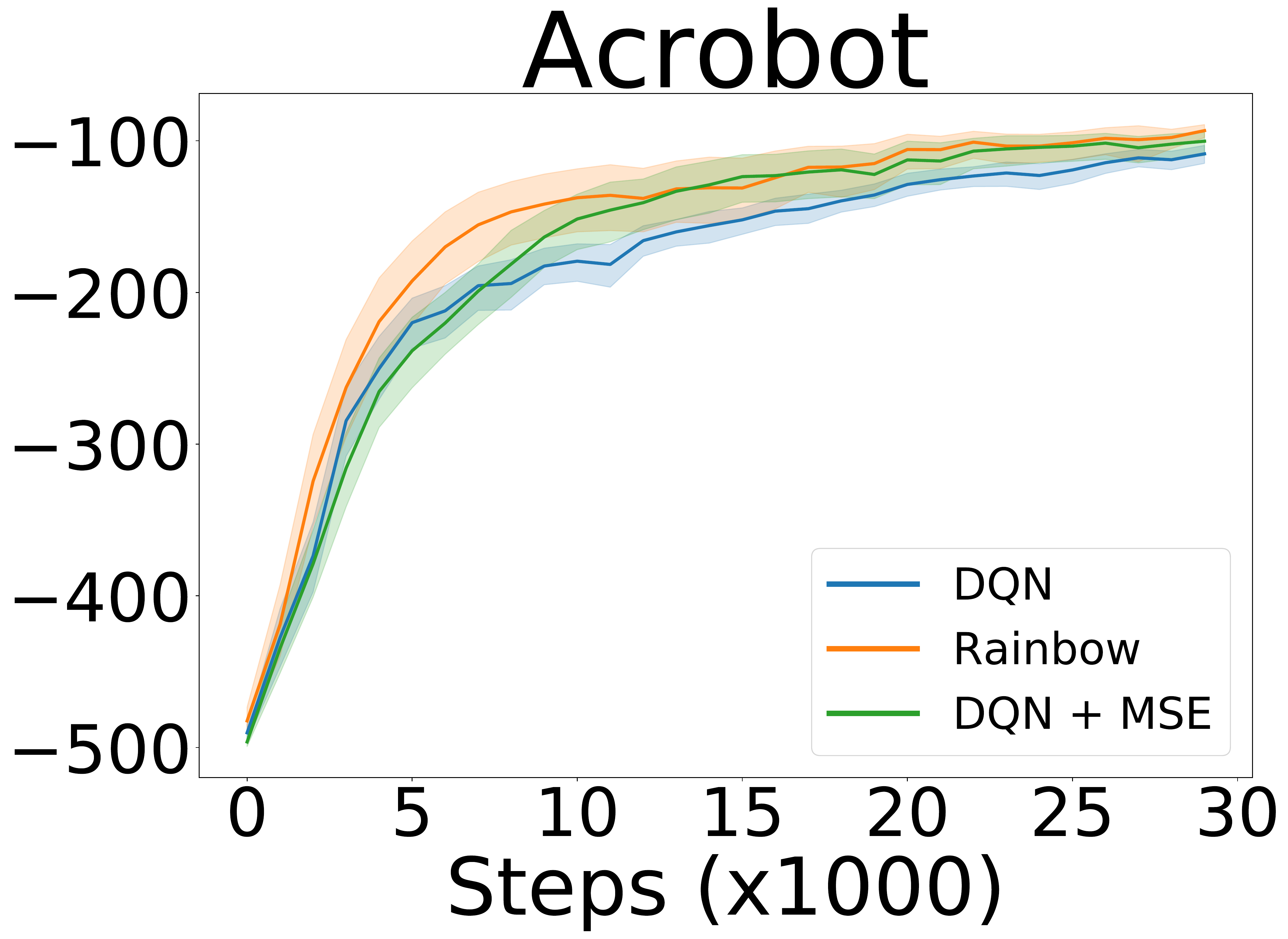}
  \includegraphics[width=0.24\textwidth]{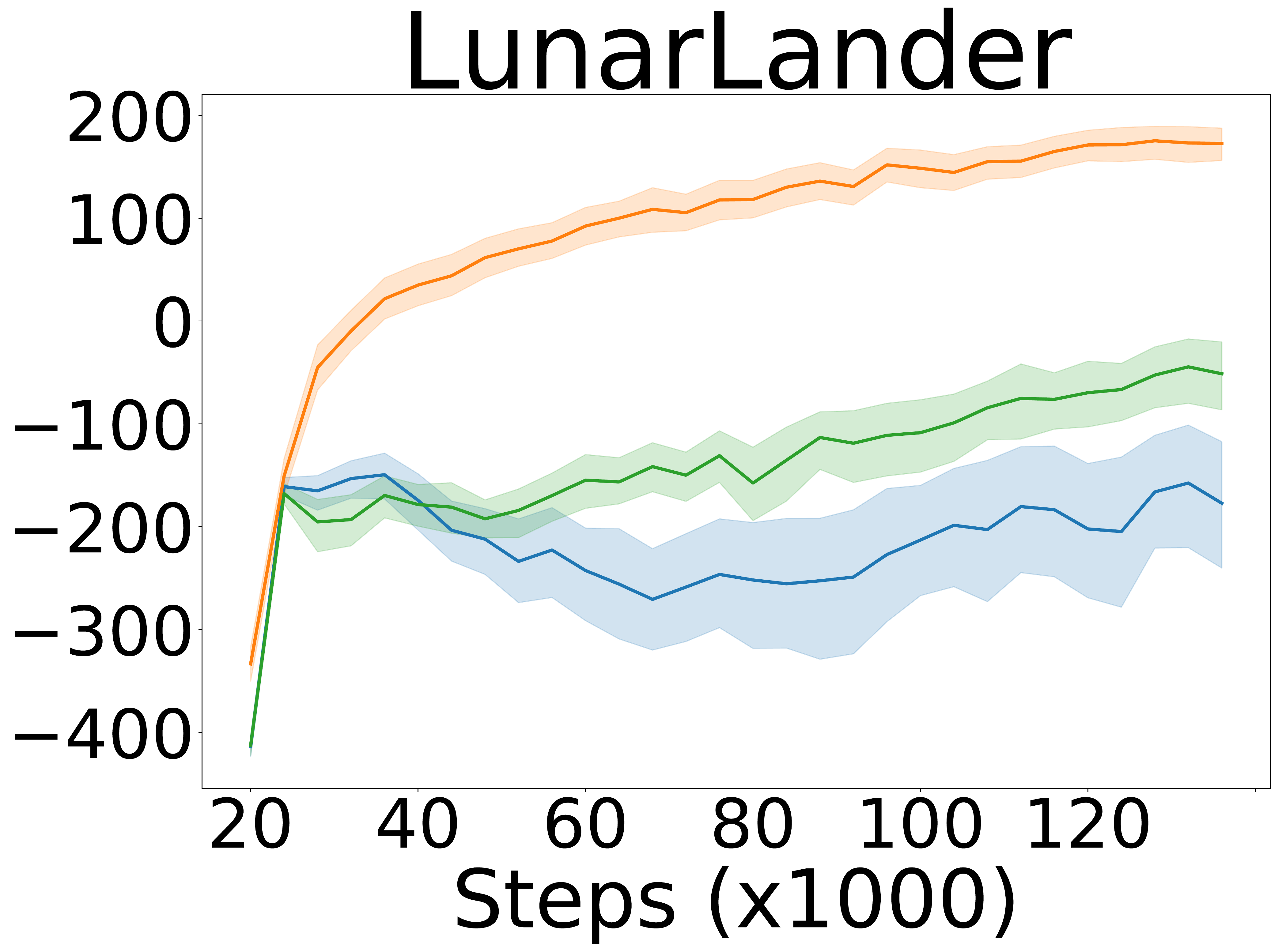}
  \includegraphics[width=0.24\textwidth]{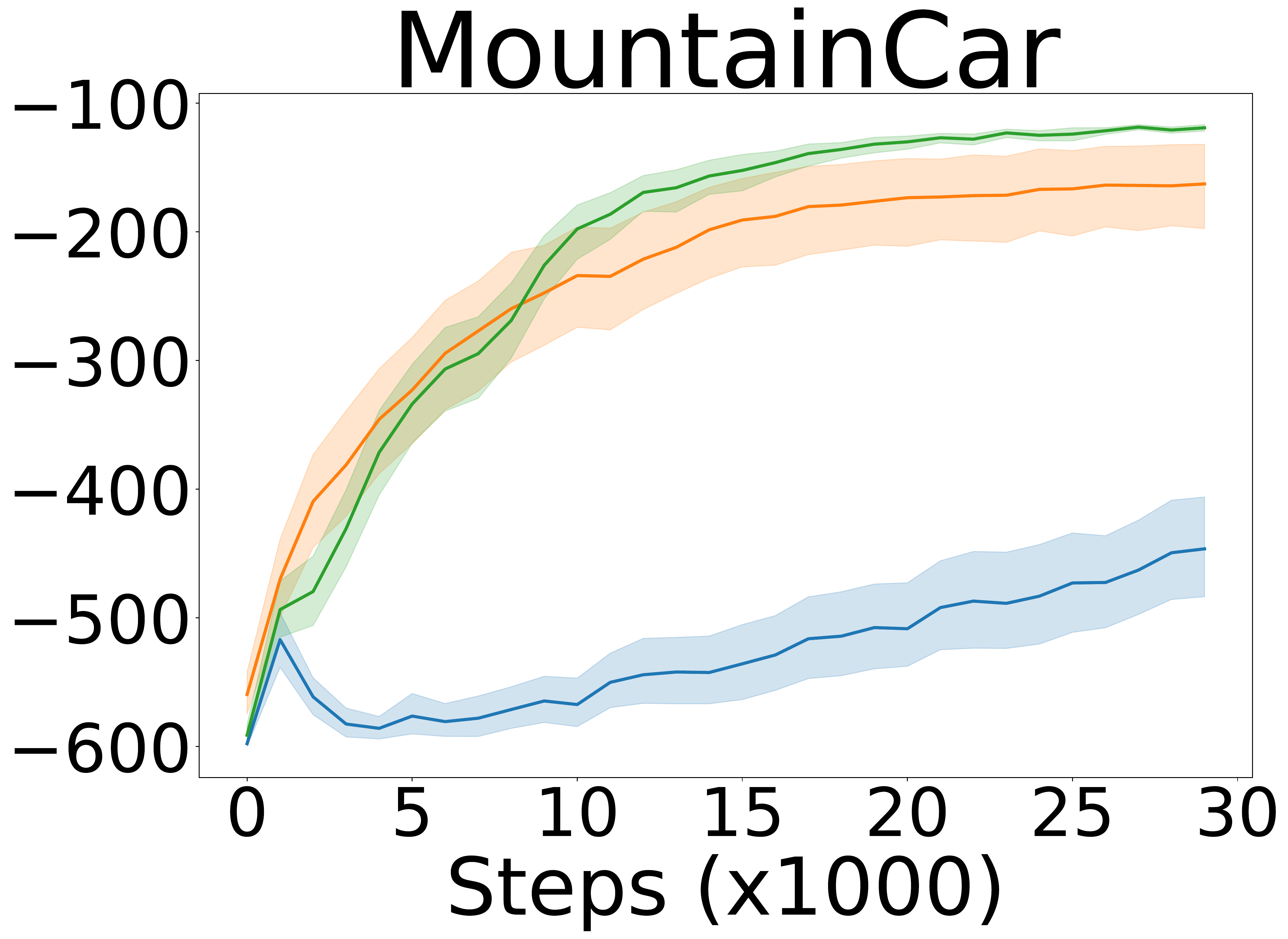}
  \includegraphics[width=0.19\textwidth]{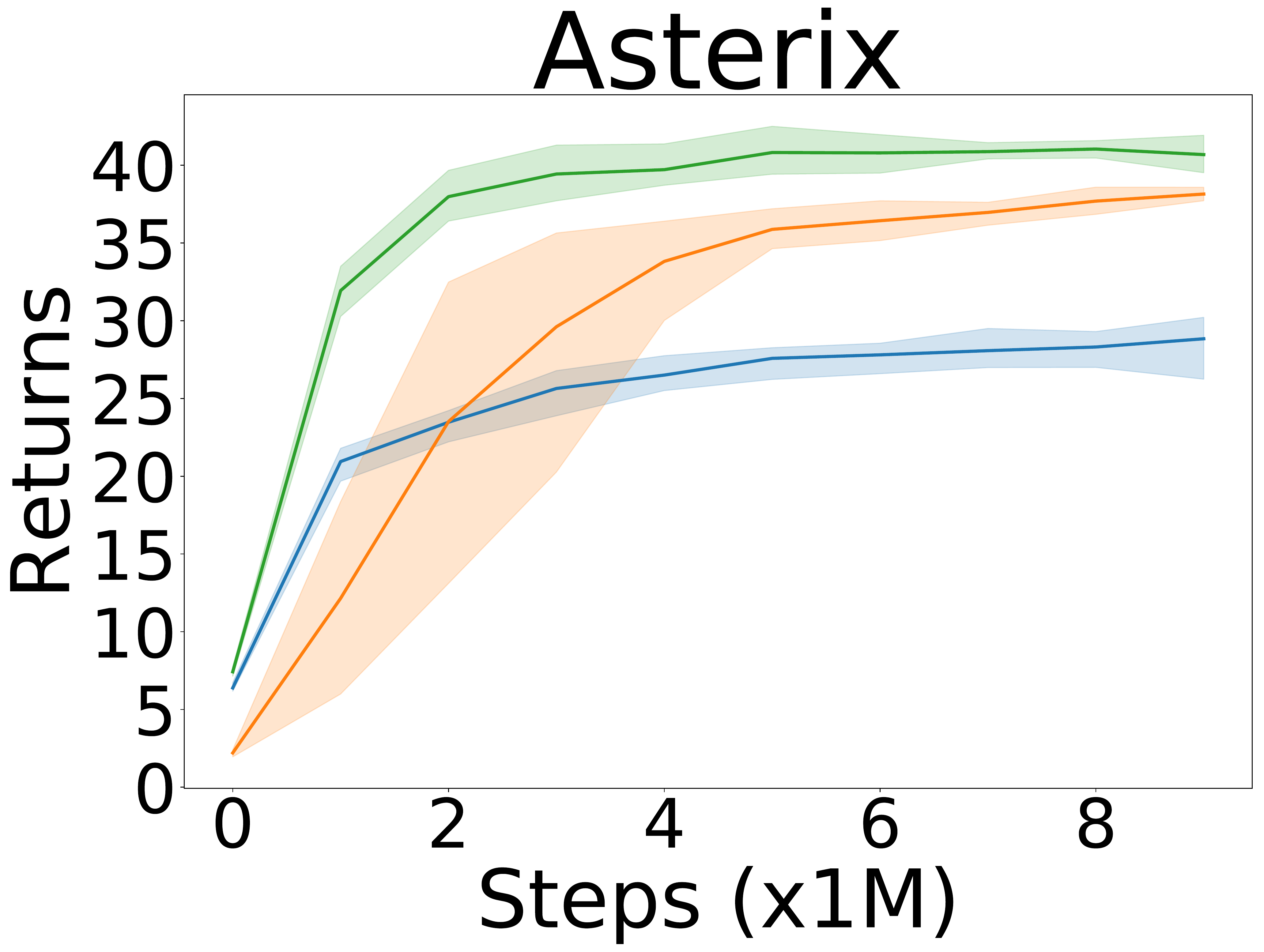}
  \includegraphics[width=0.19\textwidth]{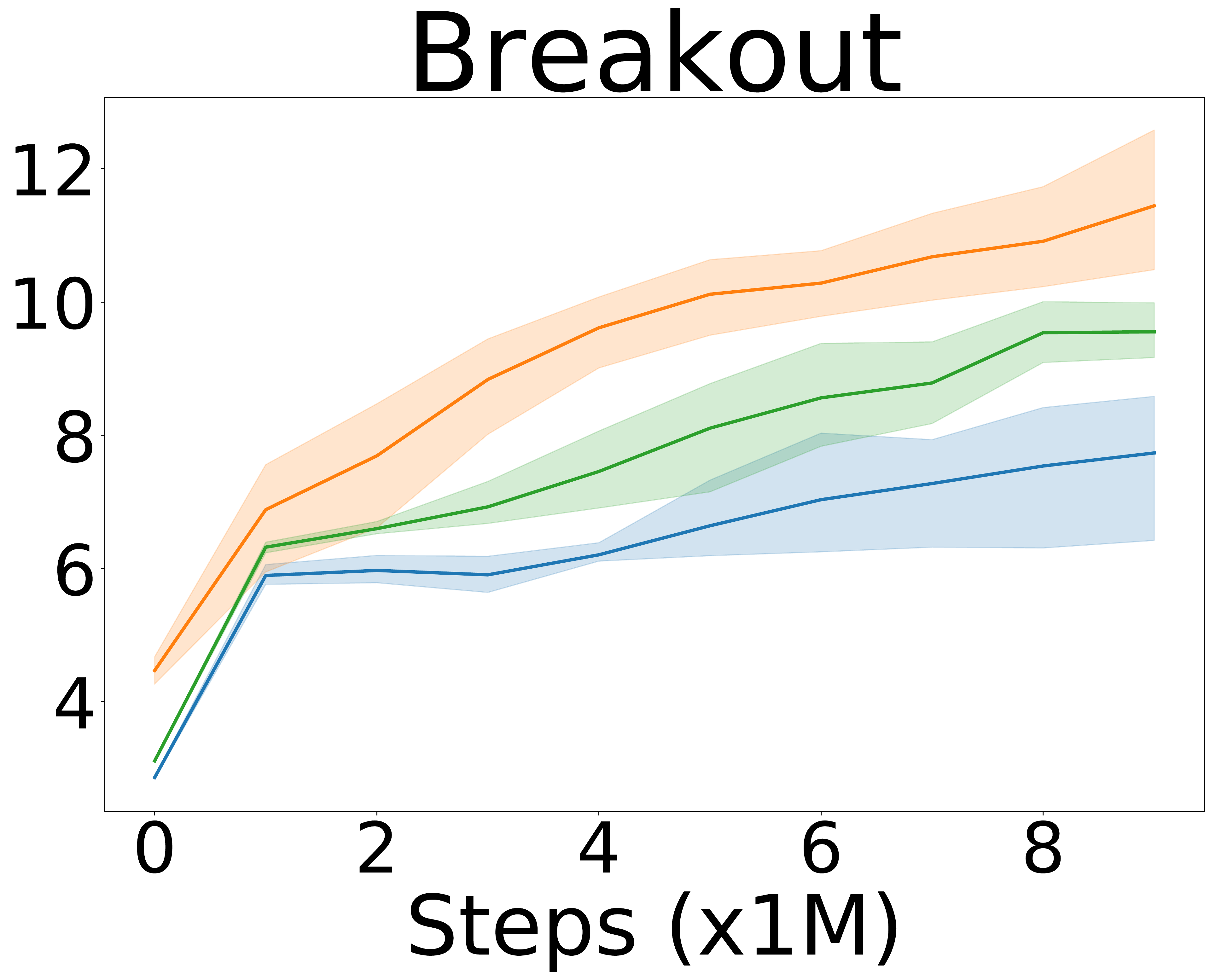}
  \includegraphics[width=0.19\textwidth]{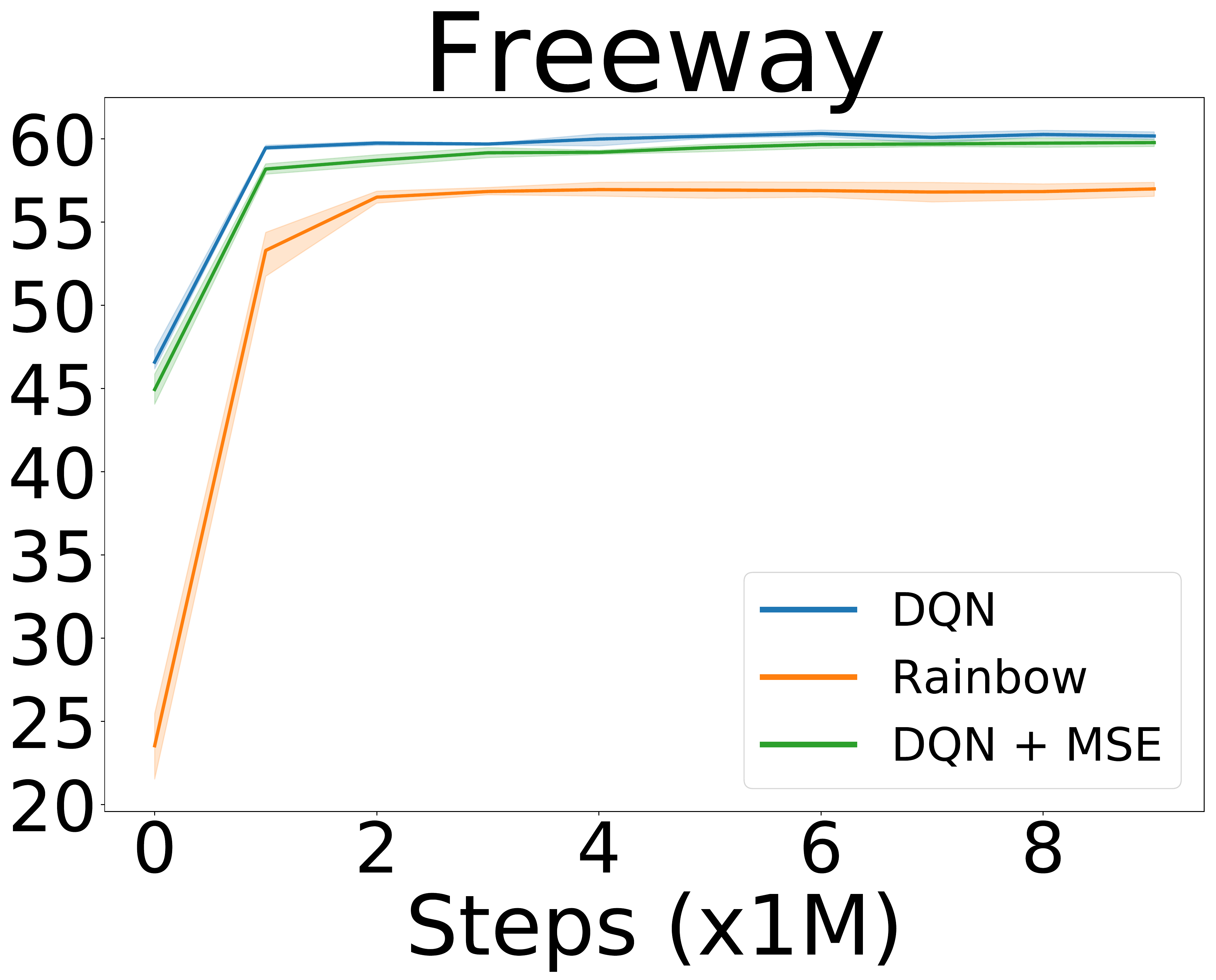}
  \includegraphics[width=0.19\textwidth]{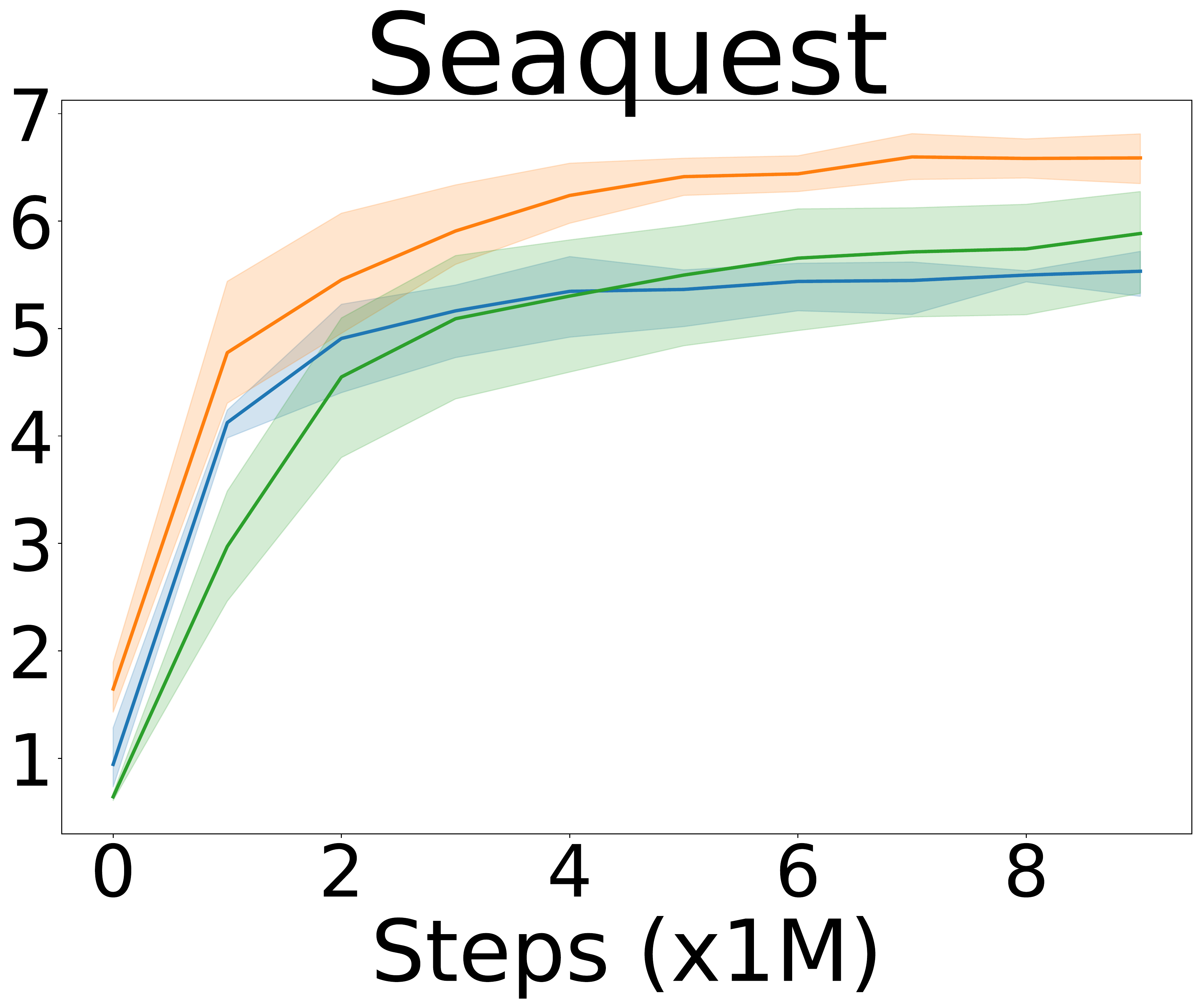}
  \includegraphics[width=0.19\textwidth]{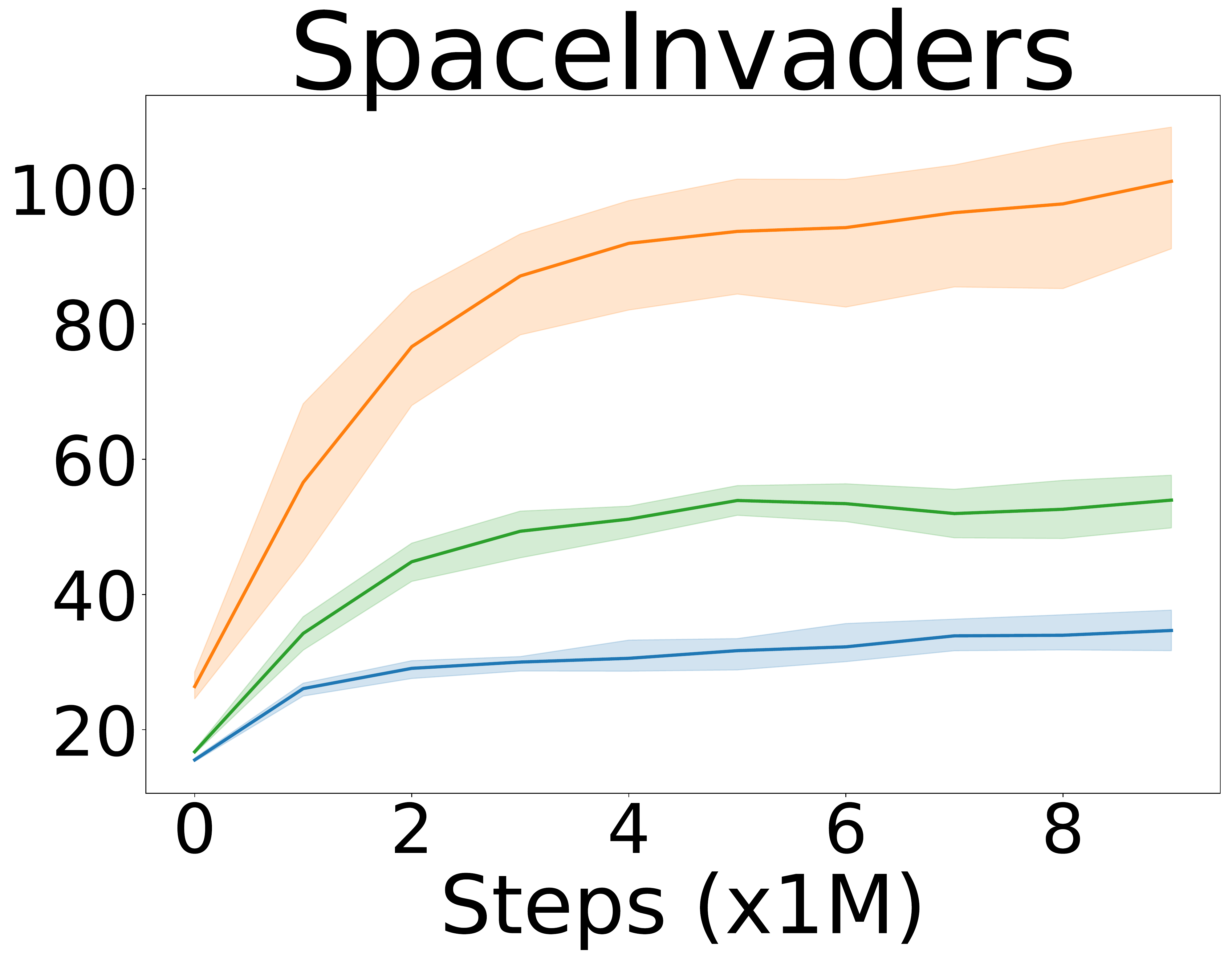}
  \caption{Evaluation of the use of the mean-squared error loss, instead of the Huber loss, in DQN.}
  \label{fig:comparisonMSE}
\end{figure*}

We first evaluated the various combinations on all the classic control and MinAtar environments and found that Adam+MSE worked best. Full results displayed in \autoref{fig:comparisonMSE}.

\subsection{Atari results}
Given that the standard practice when training DQN is to use RMSPRop with the Huber loss, yet our results on the classic control and MinAtar environments suggest that MSE would work better, we performed a thorough comparison of the various combinations, which we present in this section.

\begin{figure}[!h]
  \centering
  \includegraphics[width=\textwidth]{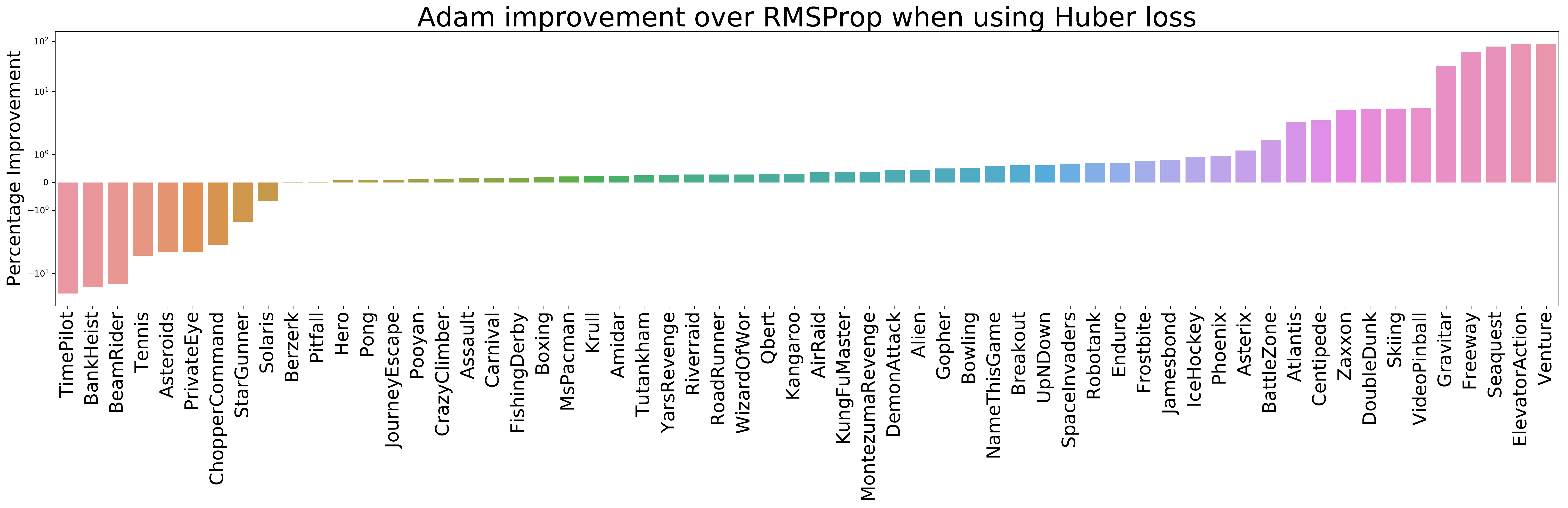}
  \caption{Comparing Adam vs RMSProp when both optimizers use the Huber loss.}
  \label{fig:AdamVsRMSPropHuber}
\end{figure}

\begin{figure}[!h]
  \centering
  \includegraphics[width=\textwidth]{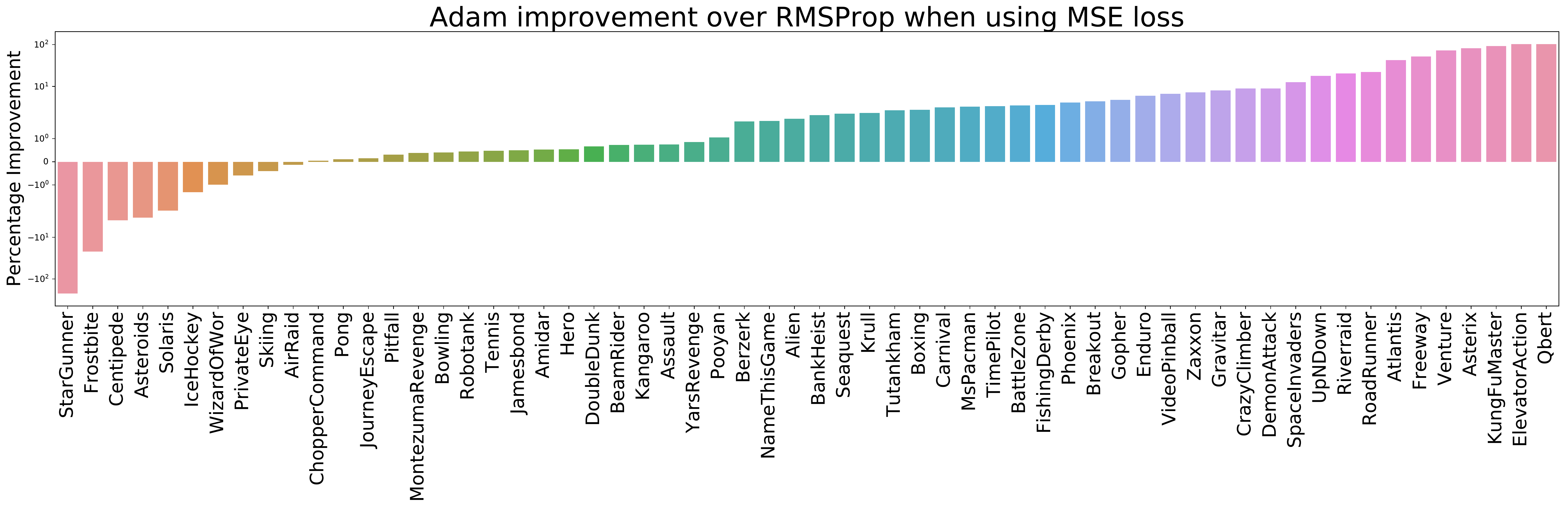}
  \caption{Comparing Adam vs RMSProp when both optimizers use the MSE loss.}
  \label{fig:AdamVsRMSPropMSE}
\end{figure}

\begin{figure}[!h]
  \centering
  \includegraphics[width=\textwidth]{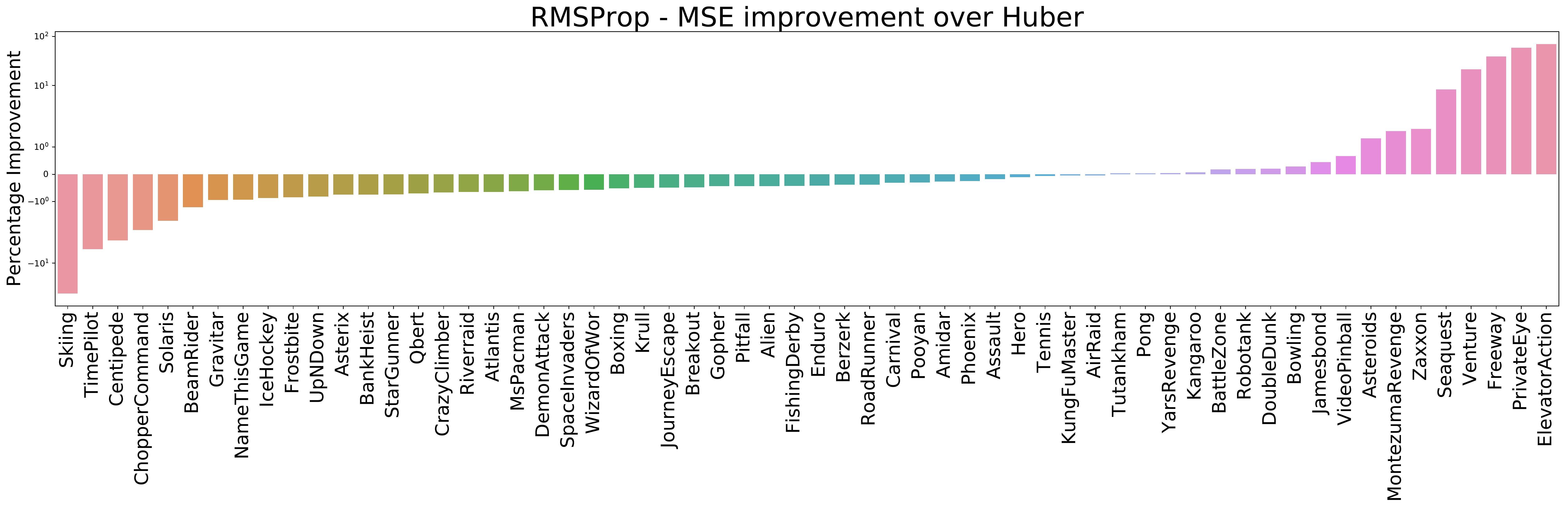}
  \caption{Comparing the MSE vs Huber loss when using the RMSProp optimizer.}
  \label{fig:MSEVsHuberRMSProp}
\end{figure}

\begin{figure}[!h]
  \centering
  \includegraphics[width=\textwidth]{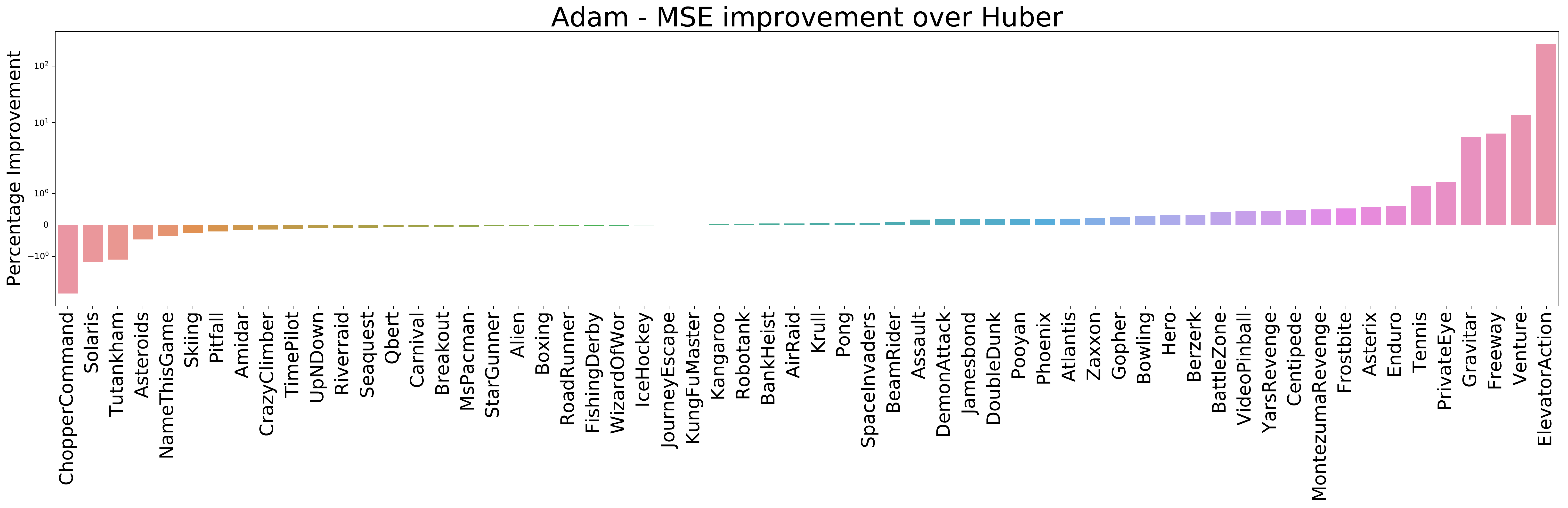}
  \caption{Comparing the MSE vs Huber loss when using the Adam optimizer.}
  \label{fig:MSEVsHuberAdam}
\end{figure}

\begin{figure*}
  \centering
  \includegraphics[width=0.45\textwidth]{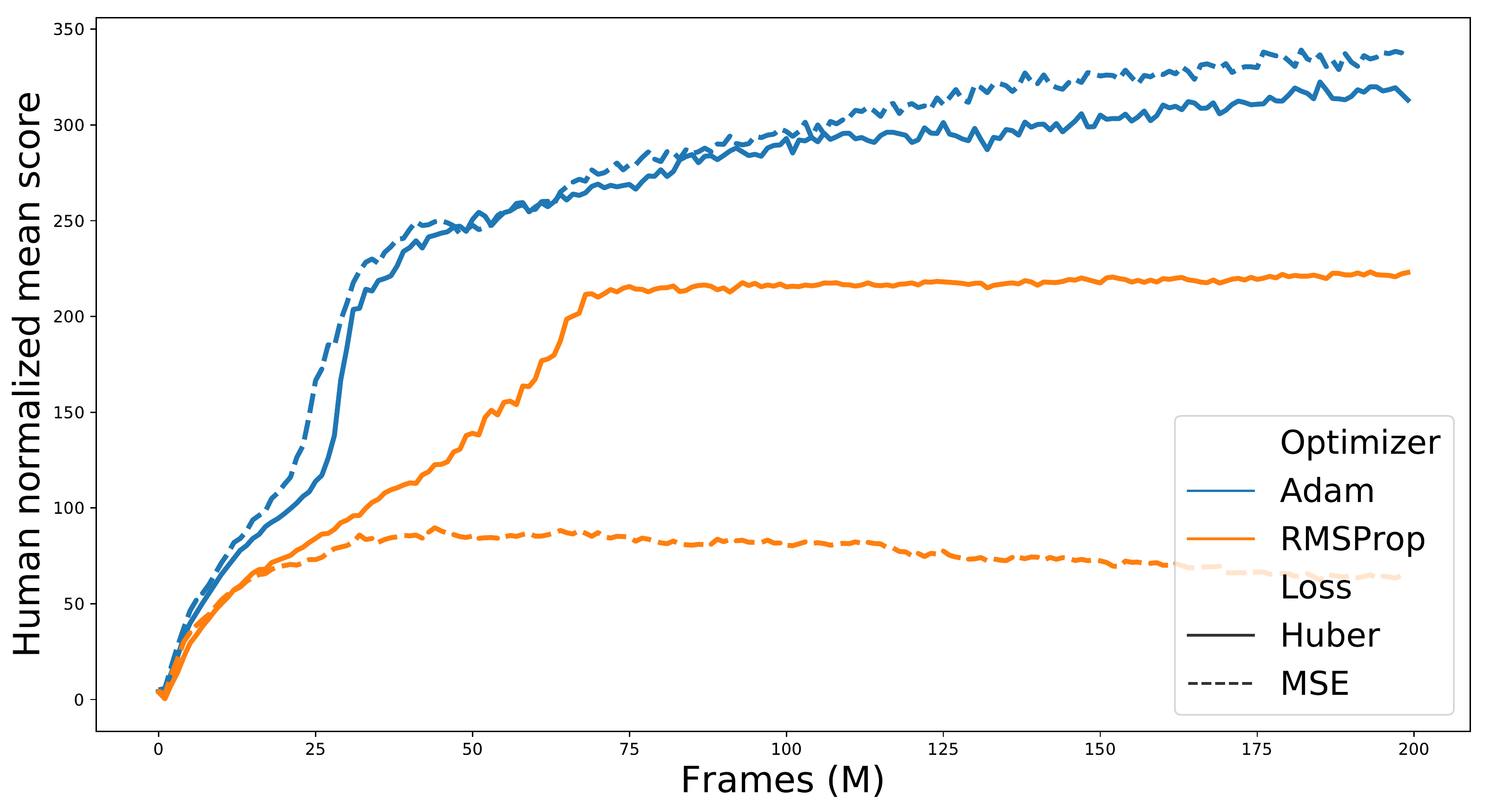}
  \includegraphics[width=0.45\textwidth]{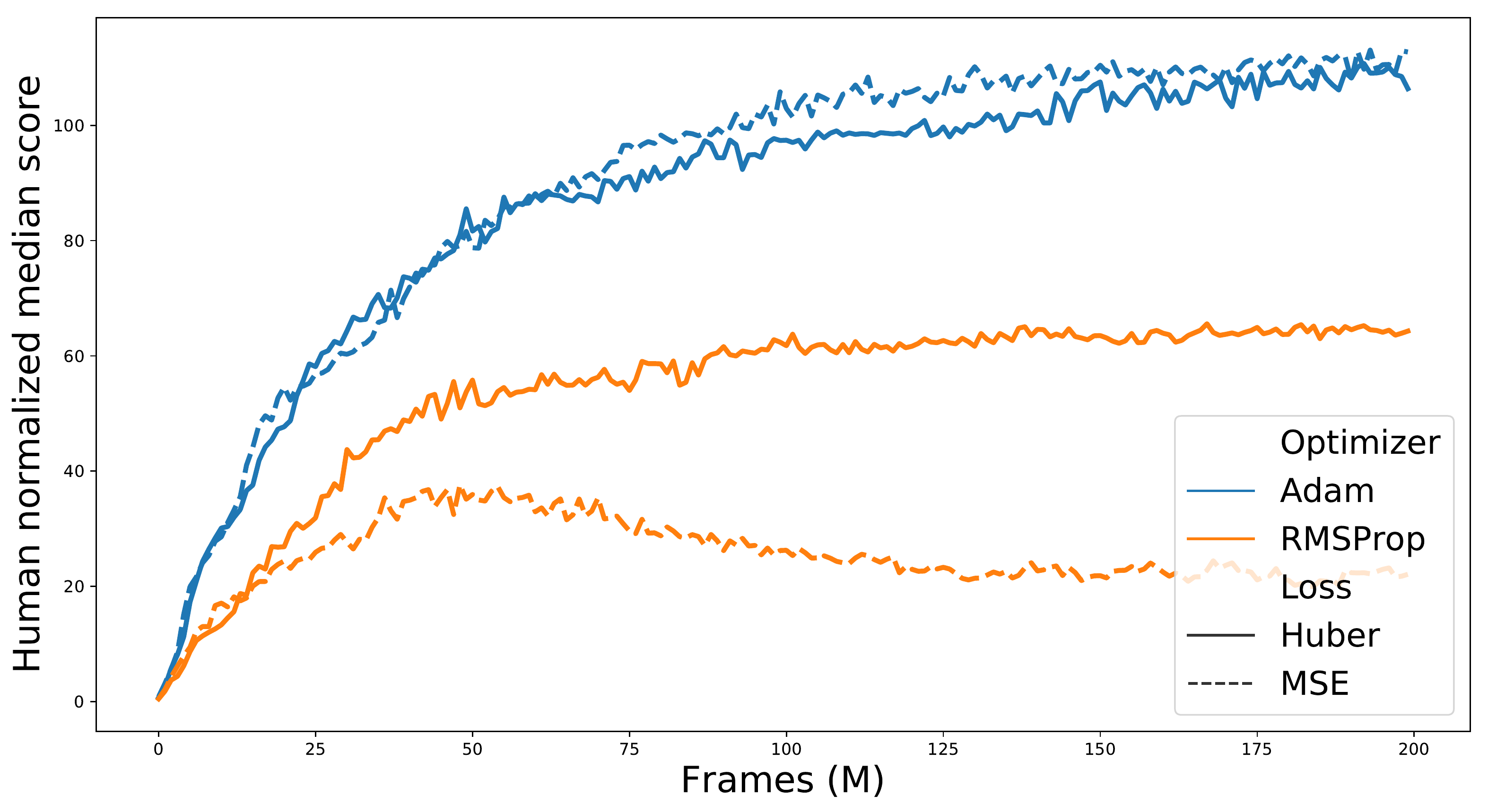}
  \caption{Comparison of the various optimizer-loss combinations with human normalized scores (mean left, median right). All results report the average of 5 independent runs.}
  \label{fig:optimizerLossComparisonsHumanNormalized}
\end{figure*}

\begin{figure*}[!t]
  \centering
  \includegraphics[width=0.8\textwidth]{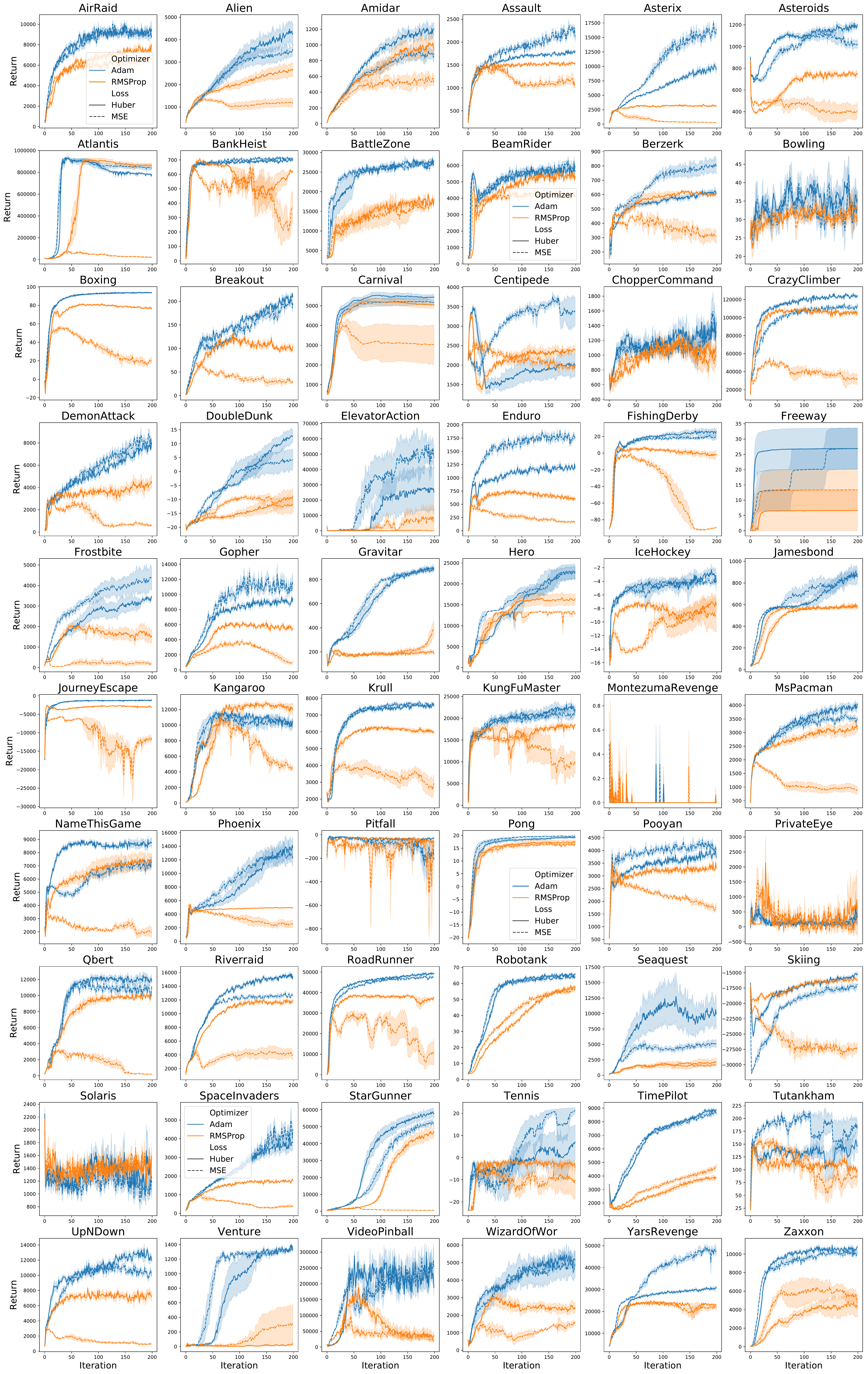}
  \caption{Comparison of using Adam (blue) versus RMSProp (orange) with Huber loss (solid) versus MSE loss (dashed) on all 60 Atari games. Each combination was run over 5 independent runs and the shaded areas represent 75\% confidence intervals.}
  \label{fig:fullAtari}
\end{figure*}

We compare Adam vs RMSProp when both use the Huber loss (\autoref{fig:AdamVsRMSPropHuber}) and MSE loss (\autoref{fig:AdamVsRMSPropMSE}); we also compare the MSE vs Huber loss when using RMSProp (\autoref{fig:MSEVsHuberRMSProp}) and using Adam (\autoref{fig:MSEVsHuberAdam}). In \autoref{fig:optimizerLossComparisonsHumanNormalized} we compare the various combinations using human normalized scores across all games, and we provide complete training curves for all games in \autoref{fig:fullAtari}. As can be seen, the best results are obtained when using Adam and the MSE loss, in contrast to the standard practice of using RMSProp and the Huber loss.

\clearpage

\clearpage

\section{Rainbow flavours, full results}
\begin{figure*}[!h]
  \centering
  \includegraphics[width=\textwidth]{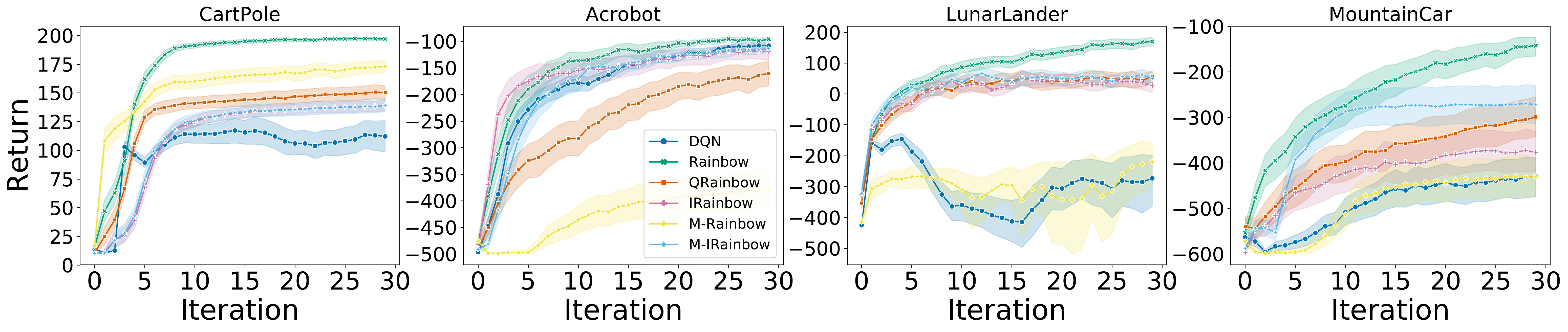}
  \includegraphics[width=\textwidth]{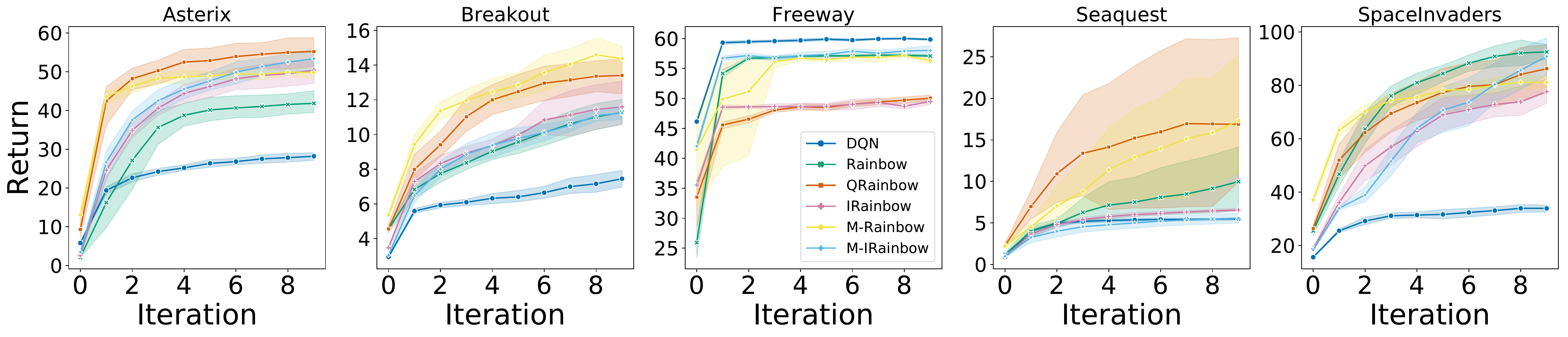}
  \caption{Comparing DQN against the various Rainbow flavours.}
  \label{fig:aggregateComparisonsFull}
\end{figure*}

In this section we present the complete results for all the environments when comparing DQN against the various Rainbow flavours (\autoref{fig:aggregateComparisonsFull}).

\end{document}